\documentclass{article}
\pdfoutput=1

\PassOptionsToPackage{numbers}{natbib}
\usepackage[final]{nips_2016}

\usepackage[utf8]{inputenc} 
\usepackage[T1]{fontenc}    
\usepackage{url}            
\usepackage{booktabs}       
\usepackage{amsfonts}       
\usepackage{nicefrac}       
\usepackage{microtype}      

\usepackage{times}
\usepackage{graphicx} 
\usepackage{subfig}
\usepackage{xcolor}

\usepackage{amsmath}
\usepackage[bookmarks=false]{hyperref}       

\newcommand\crule[3][black]{\textcolor{#1}{\rule{#2}{#3}}}
\definecolor{plot-linf}{HTML}{8172b2}
\definecolor{plot-l2}{HTML}{ccb974}
\definecolor{plot-l1}{HTML}{64b5cd}
\definecolor{plot-algo}{HTML}{c44e52}
\definecolor{plot-policy}{HTML}{4c72b0}
\definecolor{plot-none}{HTML}{55a868}

\newcommand{\sref}[1]{Sec. \ref{#1}}
\newcommand{\figref}[1]{Fig. \ref{#1}}

\begin{document} 

\title{Adversarial Attacks on Neural Network Policies}

\author{
  Sandy Huang$^\dagger$, Nicolas Papernot$^\ddagger$, Ian Goodfellow$^\S$, Yan Duan$^\dagger$$^\S$, Pieter Abbeel$^\dagger$$^\S$ \\
  $^\dagger$ University of California, Berkeley, Department of Electrical Engineering and Computer Sciences \\
  $^\ddagger$ Pennsylvania State University, School of Electrical Engineering and Computer Science \\
  $^\S$ OpenAI
}

\maketitle

\begin{abstract}
Machine learning classifiers are known to be vulnerable to inputs
maliciously constructed by adversaries to force misclassification.
Such adversarial examples have been extensively studied in the
context of computer vision applications. In this work, we show
adversarial attacks are also effective when targeting neural network
policies in reinforcement learning. Specifically, we show existing
adversarial example crafting techniques can be used to significantly
degrade test-time performance of trained policies. Our threat
model considers adversaries capable of introducing small perturbations
to the raw input of the policy. We characterize the degree of
vulnerability across tasks and training algorithms, for a subclass of 
adversarial-example attacks in white-box and black-box settings.
Regardless of the learned task or training algorithm, we observe a 
significant drop in performance, even with small adversarial 
perturbations that do not interfere with human perception.
Videos are available
at \url{http://rll.berkeley.edu/adversarial}.
\end{abstract}

\section{Introduction}

Recent advances in deep learning and deep reinforcement learning (RL) have made it possible to learn end-to-end policies that map directly from raw inputs (e.g., images) to a distribution over actions to take. Deep RL algorithms have trained policies that achieve superhuman performance on Atari games~\cite{Mnih_2013,Schulman_2015,Mnih_2016} and Go~\cite{Silver_2016}, perform complex robotic manipulation skills~\cite{Levine_2016}, learn locomotion tasks~\cite{Schulman_2015,Lillicrap_2016}, and drive in the real world~\cite{Bojarski_2016}.

These policies are parametrized by neural networks, which have
been shown to be vulnerable to adversarial attacks in supervised learning settings. For example, for convolutional neural networks trained to classify images, perturbations added to the input image can cause the network to classify the adversarial image incorrectly, while the two images remain essentially indistinguishable to humans~\cite{szegedy2013intriguing}.
In this work, we investigate whether such adversarial examples affect neural network \emph{policies}, which are trained with deep RL. We consider a fully trained policy at test time, and allow the adversary to make limited changes to the raw input perceived from the environment before it is passed to the policy.

Unlike supervised learning applications, where a fixed dataset of training examples is processed during learning, in reinforcement learning these examples are gathered throughout the training process. In other words, the algorithm used to train a policy, and even the random initialization of the policy network's weights, affects the states and actions encountered during training. Policies trained to do the same task could conceivably be significantly different (e.g., in terms of the high-level features they extract from the raw input), depending on how they were initialized and trained. Thus, particular learning algorithms may result in policies more resistant to adversarial attacks.
One could also imagine that the differences between supervised learning and
reinforcement learning might prevent an adversary from mounting a successful
attack in the black-box scenario, where the attacker does not have access to
the target policy network.

\begin{figure*}[t!]
\centering
\begin{tabular}{c}
\subfloat{\includegraphics[width = 5.3in,natwidth=864,natheight=317]{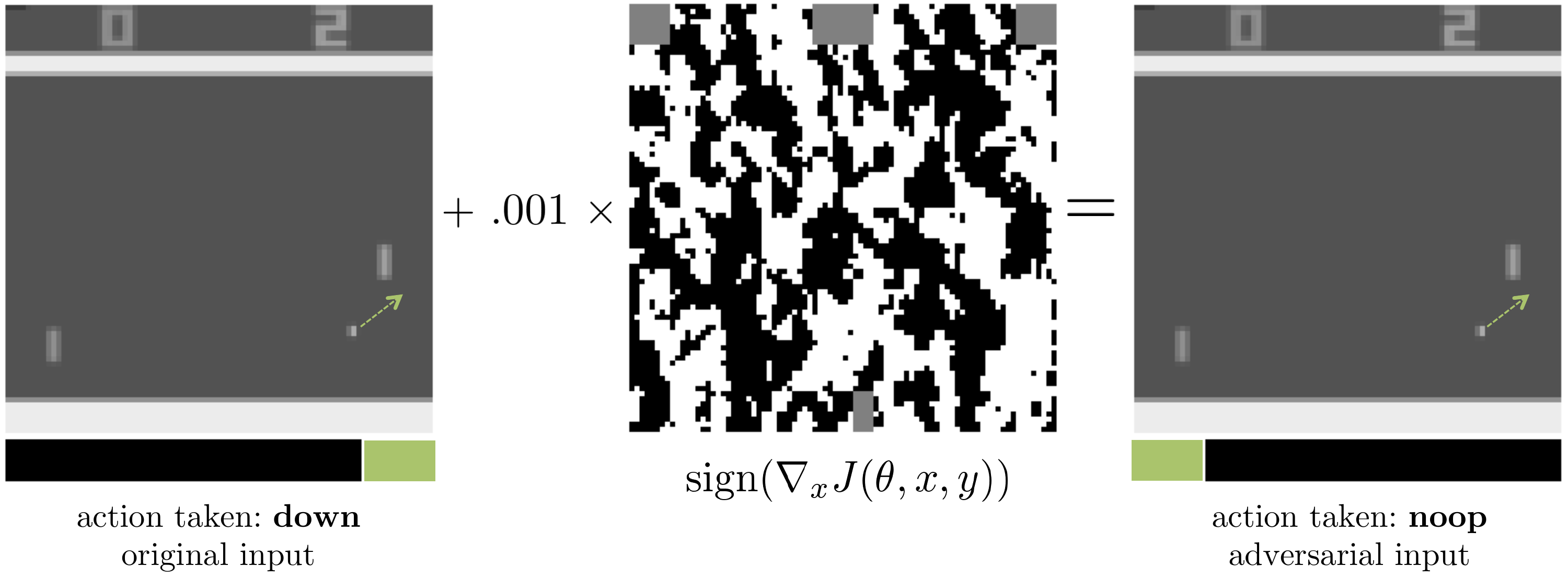}} \\ \hline
\subfloat{\includegraphics[width = 5.3in,natwidth=864,natheight=318]{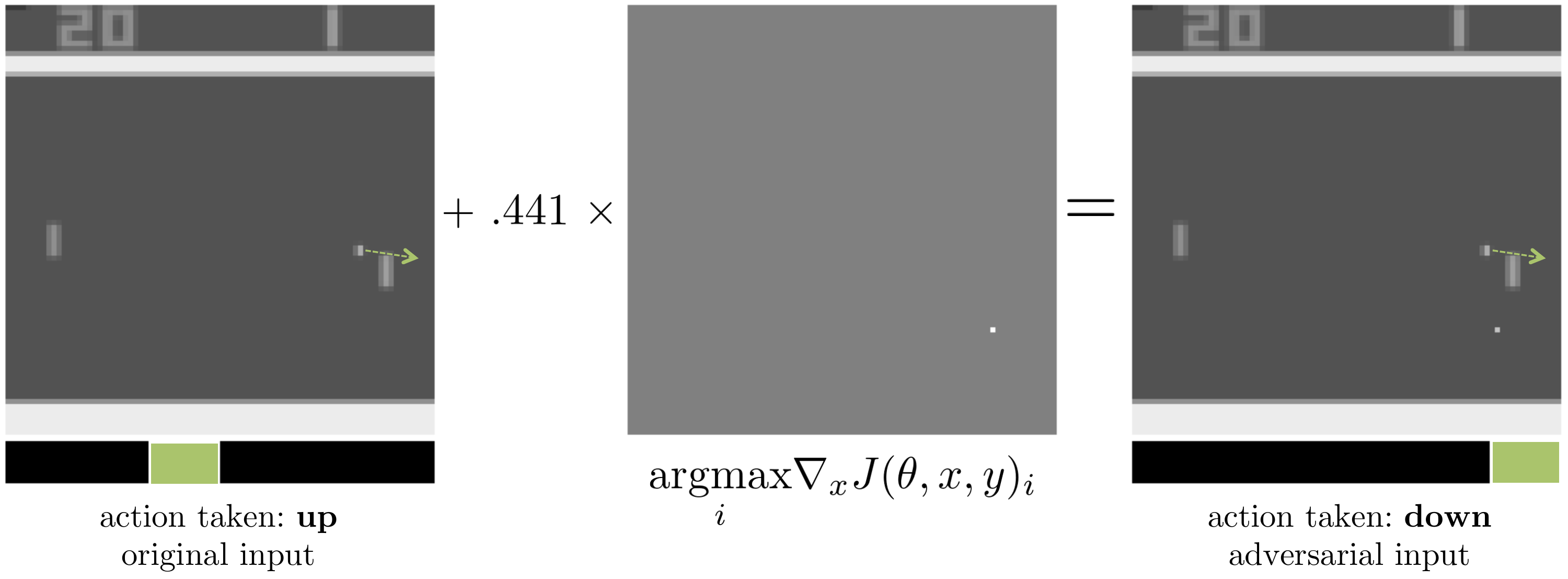}}
\end{tabular}
\caption{Two approaches for generating adversarial examples, applied to a policy trained using DQN~\cite{Mnih_2013} to play Pong. The dotted arrow starts from the ball and denotes the direction it is traveling in, and the green rectangle highlights the action that maximizes the Q-value, for the given input. In both cases, the policy chooses a good action given the original input, but the adversarial perturbation results in missing the ball and losing the point. \textbf{Top:} This adversarial example is computed using the fast gradient sign method (FGSM)~\cite{Goodfellow_2015} with an $\ell_\infty$-norm constraint on the adversarial perturbation; the adversarial input is equivalent to the original input when converted to 8-bit image encodings, but is still able to harm performance. \textbf{Bottom:} FGSM with an $\ell_1$-norm constraint; the optimal perturbation is to create a ``fake'' ball lower than the position of the actual ball.}
\label{fig:example}
\end{figure*}

Our main contribution is to characterize how the effectiveness of adversarial examples is impacted by two factors: the deep RL algorithm used to learn the policy, and whether the adversary has access to the policy network itself (white-box vs. black-box).
We first analyze three types of white-box attacks on four Atari games trained with three deep reinforcement learning algorithms (DQN~\cite{Mnih_2013}, TRPO~\cite{Schulman_2015}, and A3C~\cite{Mnih_2016}). We show that across the board, these trained policies are vulnerable to adversarial examples. However, policies trained with TRPO and A3C seem to be more resistant to adversarial attacks. \figref{fig:example} shows two examples of adversarial attacks on a Pong policy trained with DQN, each at a specific time step during test-time execution. 

Second, we explore black-box attacks on these same policies, where we assume the adversary has access to the training environment (e.g., the simulator) but not the random initialization of the target policy, and additionally may not know what the learning algorithm is.
In the context of computer vision, Szegedy et al. \cite{szegedy2013intriguing} observed the
{\em transferability property}: an adversarial example designed to
be misclassified by one model is often misclassified by other models trained to
solve the same task.
We observe that the cross-dataset transferability property also holds in 
reinforcement learning applications, in the sense that an adversarial example
designed to interfere with the operation of one policy interferes with the operation
of another policy, so long as both policies have
been trained to solve the same task.
Specifically, we observe that adversarial examples transfer
between models trained using different trajectory rollouts
and between models trained with different training algorithms.

\section{Related Work}

Adversarial machine learning~\cite{barreno2006can}, and
more generally the security and privacy of
machine learning~\cite{papernot2016towards}, encompasses a
line of work that seeks to understand
the behavior of models and learning algorithms
in the presence of adversaries.
Such malicious individuals can target machine learning systems either
during learning by tampering with the training data~\cite{biggio2012poisoning},
or during inference by manipulating inputs on which
the model is making predictions~\cite{biggio2013evasion}.
Among the perturbations crafted at test time, a class of adversarial
inputs known as \emph{adversarial examples} was
introduced by~\cite{szegedy2013intriguing}. This first demonstration of
the vulnerability of --- then state-of-the-art --- architectures
to perturbations indistinguishable to the human eye led
to a series of follow-up work showing that perturbations
could be produced with minimal computing resources~\cite{Goodfellow_2015} and/or
with access to the model label predictions only
(thus enabling black-box attacks)~\cite{papernot2016practical}, and that these perturbations
can also be applied to physical objects~\cite{Kurakin_2016,sharif2016accessorize}.

Most work on adversarial examples so far has studied their
effect on supervised learning algorithms.
A recent technical report studied the scenario of an adversary
interfering with the training of an agent, with the intent
of preventing the agent from learning anything meaningful~\cite{behzadan2017vulnerability}.
Our work is the first to study the ability of an adversary
to interfere with the operation of an RL agent by presenting
adversarial examples at test time.

\section{Preliminaries}

In this section, we describe technical background on adversarial example
crafting and deep reinforcement learning, which are used 
throughout the paper.

\subsection{Adversarial Example Crafting with the Fast Gradient Sign Method}
\label{sec:fgsm}

Techniques for crafting adversarial examples generally focus
on maximizing some measure of harm caused by an adversarial
perturbation, constrained by some limit on the size of the
perturbation intended to make it less noticeable to a human
observer.
A range of crafting techniques exist, allowing the attacker
to choose an attack that makes the right tradeoff between
computational cost and probability of success.

For exploratory research purposes, it is common to use
a computationally cheap method of generating adversarial
perturbations, even if this reduces the attack success rate
somewhat.
We therefore use the Fast Gradient Sign Method (FGSM)~\cite{Goodfellow_2015}, an existing method for efficiently generating adversarial examples in the context of computer vision classification. The FGSM is fast because it makes a linear approximation of a deep model and solves the maximization problem
analytically, in closed form.
Despite this approximation, it is still able to reliably fool
many classifiers for computer vision problems, because deep models
often learn piece-wise linear functions with surprisingly large
pieces.

FGSM focuses on adversarial perturbations where each pixel of the input image is changed by no more than $\epsilon$.
Given a linear function $g(x) = w^\top x$, the optimal adversarial perturbation $\eta$ that satisfies $\| \eta \|_\infty < \epsilon$ is
\begin{equation}
\eta = \epsilon \text{ } \text{sign}(w),
\end{equation}
since this perturbation maximizes the change in output for the adversarial example $\tilde{x}$, $g(\tilde{x}) = w^\top x + w^\top \eta$.

Given an image classification network
with parameters $\theta$ and loss $J(\theta, x, y)$, where $x$ is an image and $y$ is a distribution over all possible class labels, linearizing the loss function around the input $x$ results in a perturbation of
\begin{equation}
\eta = \epsilon \text{ } \text{sign}(\nabla_x J(\theta, x, y)).
\end{equation}

\subsection{Deep Reinforcement Learning}
Reinforcement learning algorithms train a policy $\pi$ to optimize the expected cumulative reward received over time. For a given state space $\mathcal{S}$ and action space $\mathcal{A}$, the policy may be a deterministic function mapping each state to an action: $\pi: \mathcal{S} \to \mathcal{A}$, or it may be a stochastic function mapping each state to a distribution over actions: $\pi: \mathcal{S} \to \Delta_{\mathcal{A}}$, where $\Delta_{\mathcal{A}}$ is the probability simplex on $\mathcal{A}$. Here, the state space may consist of images or low-dimensional state representations.
We choose to represent $\pi$ by a function parametrized by $\theta$, for instance $\theta$ may be a weighting on features of the state~\cite{Abbeel_2004}. In the case of deep reinforcement learning, $\theta$ are the weights of a neural network.
Over the past few years, a large number of algorithms for deep RL have been proposed, including deep Q-networks (DQN)~\cite{Mnih_2013}, trust region policy optimization (TRPO)~\cite{Schulman_2015}, and asynchronous advantage actor-critic (A3C)~\cite{Mnih_2016}. We compare the effectiveness of adversarial examples on feed-forward policies trained with each of these three algorithms.

\subsubsection{Deep Q-Networks}
\label{sec:dqn}
Instead of modeling the policy directly, a DQN~\cite{Mnih_2013} approximately computes, for each state, the Q-values for the available actions to take in that state. The Q-value $Q^*(s,a)$ for a state $s$ and action $a$ is the expected cumulative discounted reward obtained by taking action $a$ in state $s$, and following the optimal policy thereafter. A DQN represents the Q-value function via a neural network trained to minimize the squared Bellman error, using a variant of Q-learning. As this is off-policy learning, it employs an $\epsilon$-greedy exploration strategy. To reduce the variance of Q-learning updates, \emph{experience replay} is used: samples are randomly drawn from a replay buffer (where all recent transitions are stored) so that they are not correlated due to time. The corresponding policy for a DQN is obtained by choosing the action with the maximum Q-value for each state, hence it is deterministic.

\subsubsection{Trust Region Policy Optimization}
TRPO~\cite{Schulman_2015} is an on-policy batch learning algorithm. At each training iteration, whole-trajectory rollouts of a stochastic policy are used to calculate the update to the policy parameters $\theta$, while controlling the change in the policy as measured by the KL divergence between the old and new policies.

\subsubsection{Asynchronous Advantage Actor-Critic}
A3C~\cite{Mnih_2016} uses asynchronous gradient descent to speed up and stabilize learning of a stochastic policy. It is based on the actor-critic approach, where the actor is a neural network policy $\pi(a | s; \theta)$ and the critic is an estimate of the value function $V(s; \theta_v)$. During learning, small batches of on-policy samples are used to update the policy. The correlation between samples is reduced due to asynchronous training, which stabilizes learning.

\section{Adversarial Attacks}
In our work, we use FGSM both as a white-box attack to compute
adversarial perturbations for a trained neural network policy 
$\pi_\theta$ whose architecture and parameters are available to the 
adversary, and as a black-box attack by computing gradients for a 
separately trained policy $\pi'_\theta$ to attack $\pi_\theta$
using adversarial example transferability~\cite{szegedy2013intriguing,Goodfellow_2015,papernot2016practical}.

\subsection{Applying FGSM to Policies}

FGSM requires calculating $\nabla_{x} J(\theta,x,y)$, the gradient of the cost function $J(\theta,x,y)$ with respect to the input $x$. In reinforcement learning settings, we assume the output $y$ is a weighting over possible actions (i.e., the policy is stochastic: $\pi_\theta: \mathcal{S} \to \Delta_{\mathcal{A}}$). When computing adversarial perturbations with FGSM for a trained policy $\pi_\theta$, we assume the action with the maximum weight in $y$ is the optimal action to take: in other words, we assume the policy performs well at the task. Thus, $J(\theta,x,y)$ is the cross-entropy loss between $y$ and the distribution that places all weight on the highest-weighted action in $y$.\footnote{Functionally, this is equivalent to a technique introduced in the context of image classification, to generate adversarial examples without access to the true class label \cite{kurakin2017adversarial}.}

Of the three learning algorithms we consider, TRPO and A3C both train stochastic policies. However, DQN produces a deterministic policy, since it always selects the action that maximizes the computed Q-value. This is problematic because it results in a gradient $\nabla_{x} J(\theta,x,y)$ of zero for almost all inputs $x$. Thus, when calculating $J(\theta,x,y)$ for policies trained with DQN, we define $y$ as a softmax of the computed Q-values (with a temperature of 1). Note that we only do this for creating adversarial examples; during test-time execution, policies trained with DQN are still deterministic.

\subsection{Choosing a Norm Constraint}
Let $\eta$ be the adversarial perturbation. In certain situations, it may be desirable to change all input features by no more than a tiny amount (i.e., constrain the $\ell_\infty$-norm of $\eta$), or it may be better to change only a small number of input features (i.e., constrain the $\ell_1$-norm of $\eta$). Thus we consider variations of FGSM that restrict the $\ell_1$- and $\ell_2$-norm of $\eta$, as well as the original version of FGSM that restricts the $\ell_\infty$-norm (\sref{sec:fgsm}).

Linearizing the cost function $J(\theta,x,y)$ around the current input $x$, the optimal perturbation for each type of norm constraint is:
\begin{equation}
\eta = \begin{cases}
    \epsilon \text{ } \text{sign}(\nabla_x J(\theta,x,y)) \hspace{7pt} \text{ for constraint } \|\eta\|_\infty \leq \epsilon \\
    \epsilon \sqrt{d}  * \frac{\nabla_x J(\theta,x,y)}{\| \nabla_x J(\theta,x,y) \|_2} \hspace{9pt} \text{ for constraint } \|\eta\|_2 \leq \| \epsilon \mathbf{1}_d \|_2 \\
    \text{maximally perturb highest-impact dimensions with budget } \epsilon d \\
        \hspace{90pt} \text{ for constraint } \|\eta\|_1 \leq \| \epsilon \mathbf{1}_d \|_1
\end{cases}
\end{equation}
where $d$ is the number of dimensions of input $x$. 

Note that the $\ell_2$-norm and $\ell_1$-norm constraints have 
$\epsilon$ adjusted to be the $\ell_2$- and $\ell_1$-norm of the 
vector $\epsilon \mathbf{1}_d$, respectively, since that is the 
amount of perturbation under the $\ell_\infty$-norm constraint.
In addition, the optimal perturbation for the $\ell_1$-norm 
constraint either maximizes or minimizes the feature value at 
dimensions $i$ of the input, ordered by decreasing $| \nabla_\theta J(\theta,x,y)_i |$. For this norm, the adversary's budget --- the total amount of perturbation the adversary is allowed to introduce 
in the input --- is $\epsilon d$.

\section{Experimental Evaluation}
We evaluate our adversarial attacks on four Atari 2600 games in the Arcade Learning Environment~\cite{Bellemare_2013}: Chopper Command, Pong, Seaquest, and Space Invaders. We choose these games to encompass a variety of interesting environments; for instance, Chopper Command and Space Invaders include multiple enemies. 

\subsection{Experimental Setup}
We trained each game with three deep reinforcement learning algorithms: A3C~\cite{Mnih_2016}, TRPO~\cite{Schulman_2015}, and DQN~\cite{Mnih_2013}.

For DQN, we use the same pre-processing and neural network architecture as in \cite{Mnih_2013}  (Appendix~\ref{sec:exp_setup}). We also use this architecture for the stochastic policies trained by A3C and TRPO. Specifically, the input to the neural network policy is a concatenation of the last 4 images, converted from RGB to luminance (Y) and resized to $84\times84$. Luminance values are rescaled to be from 0 to 1. The output of the policy is a distribution over possible actions.

For each game and training algorithm, we train five policies starting from different random initializations. For our experiments, we focus on the top-performing trained policies, which we define as all policies that perform within $80\%$ of the maximum score for the last ten training iterations. We cap the number of policies at three for each game and training algorithm. Certain combinations (e.g., Seaquest with A3C) had only one policy meet these requirements.

In order to reduce the variance of our experimental results, the average return for each result reported is the average cumulative reward across ten rollouts of the target policy, without discounting rewards.

\subsection{Vulnerability to White-Box Attacks}
First, we are interested in how vulnerable neural network policies are to white-box adversarial-example attacks, and how this is affected by the type of adversarial perturbation and by how the policy is trained. If these attacks are effective, even small adversarial perturbations (i.e., small $\epsilon$ for FGSM) will be able to significantly lower the performance of the target trained network, as observed in~\cite{Goodfellow_2015} for image classifiers. We evaluate multiple settings of $\epsilon$ across all four games and three training algorithms, for the three types of norm-constraints for FGSM. 

\subsubsection{Observations}
\begin{figure*}[t!]
\centering
\begin{tabular}{cccc}
\subfloat{\includegraphics[width = 1.33in,natwidth=529,natheight=406]{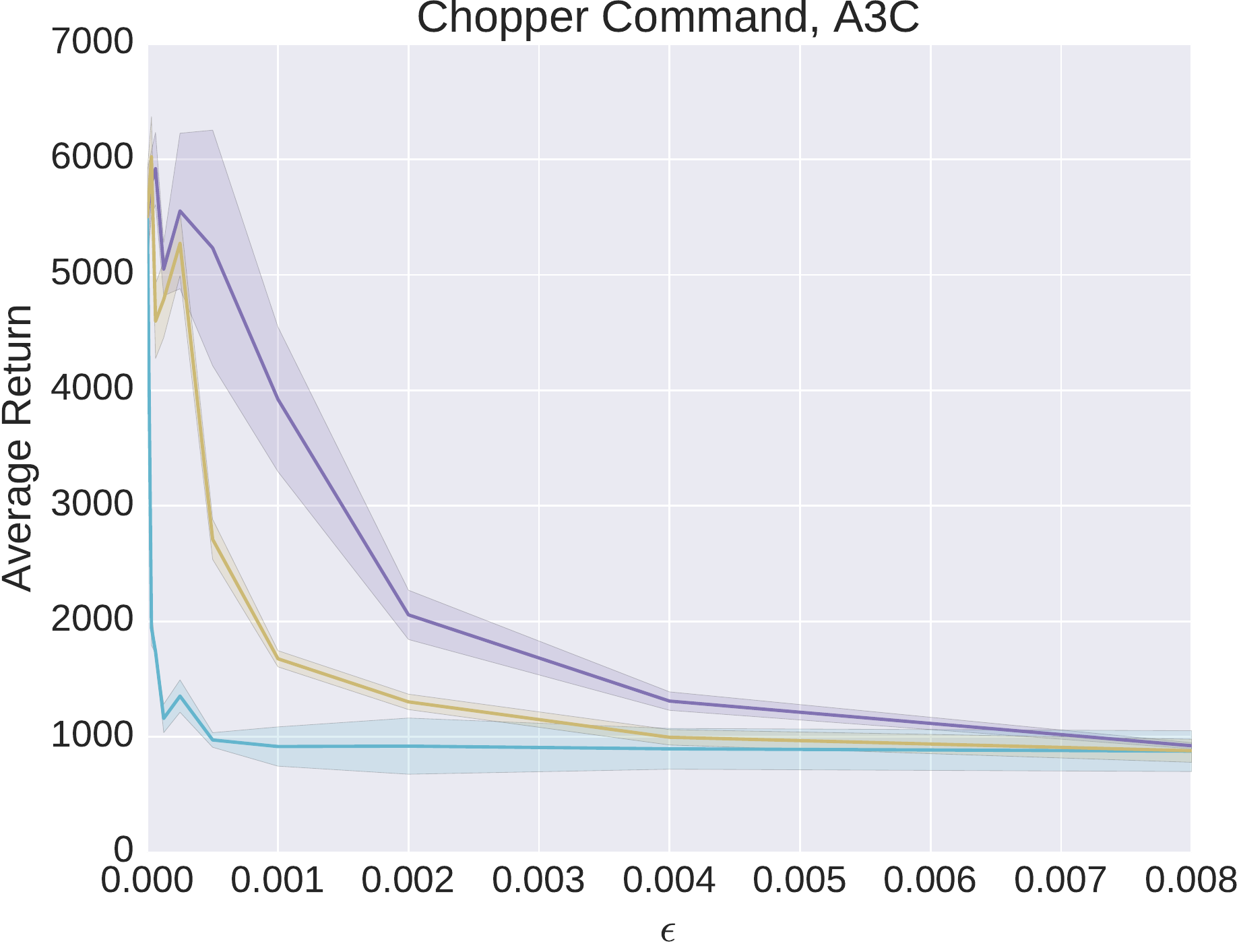}} &
\subfloat{\includegraphics[width = 1.33in,natwidth=520,natheight=406]{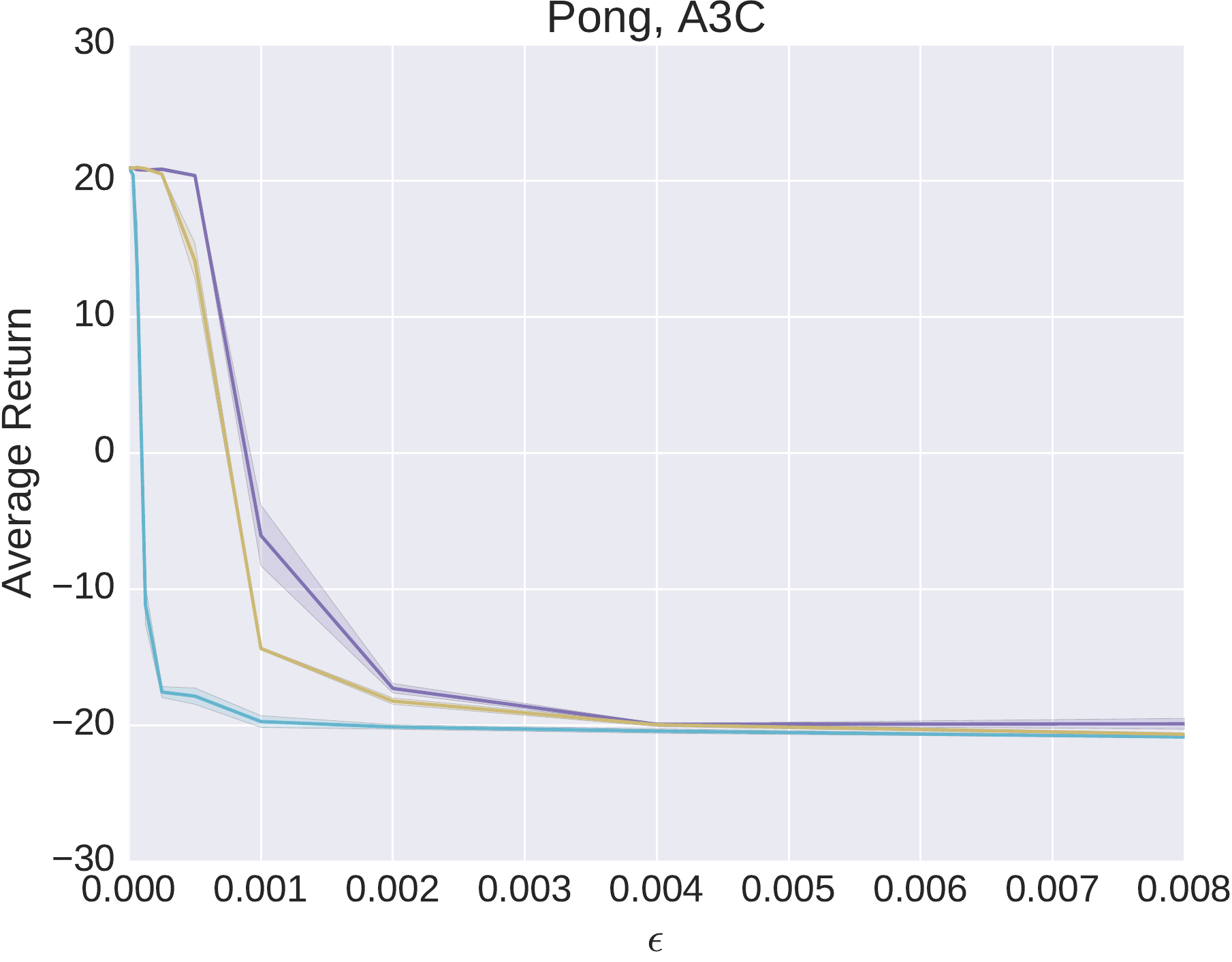}} &
\subfloat{\includegraphics[width = 1.33in,natwidth=529,natheight=406]{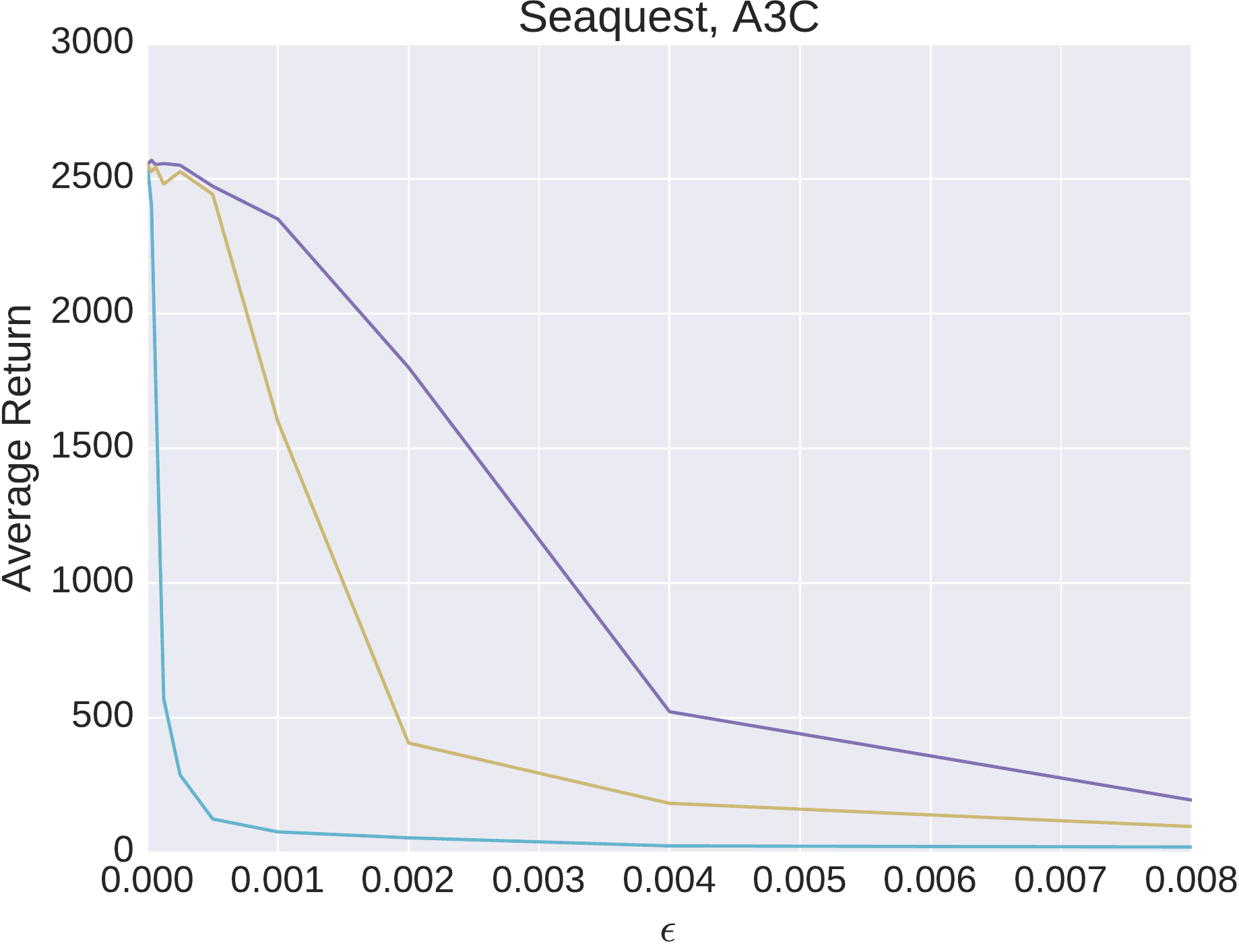}} &
\subfloat{\includegraphics[width = 1.33in,natwidth=529,natheight=406]{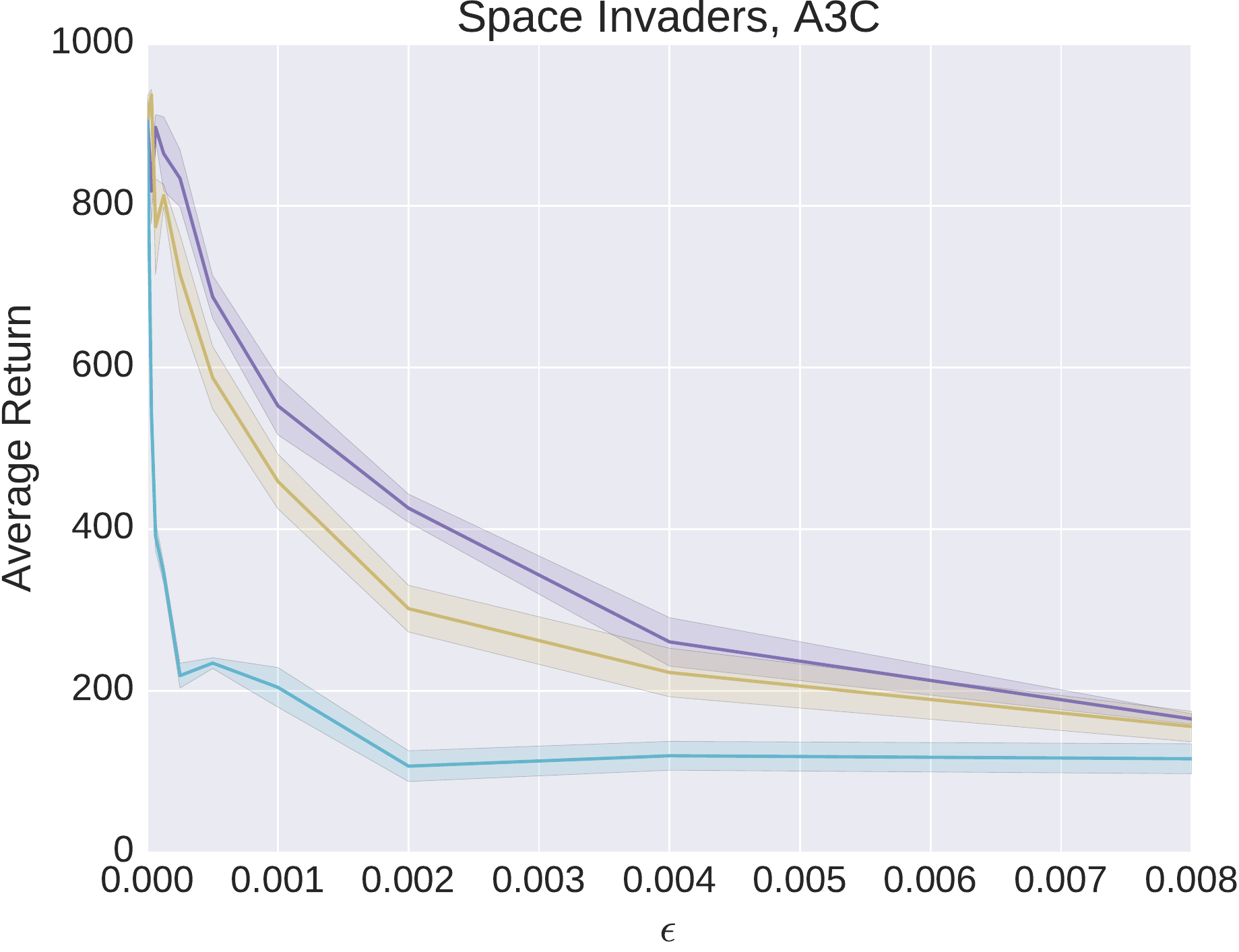}} \\
\addlinespace[-2ex]
\subfloat{\includegraphics[width = 1.33in,natwidth=529,natheight=406]{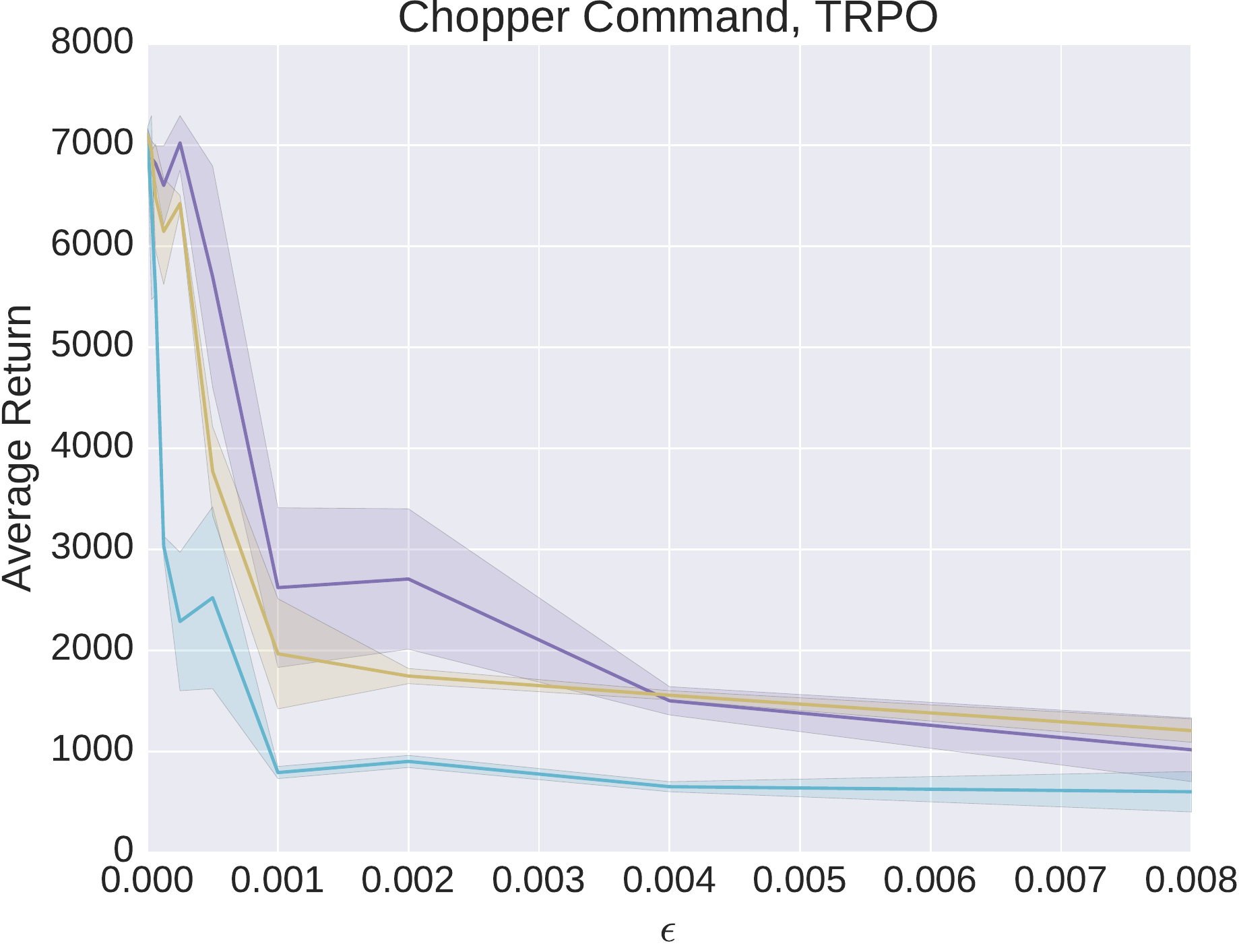}} &
\subfloat{\includegraphics[width = 1.33in,natwidth=520,natheight=406]{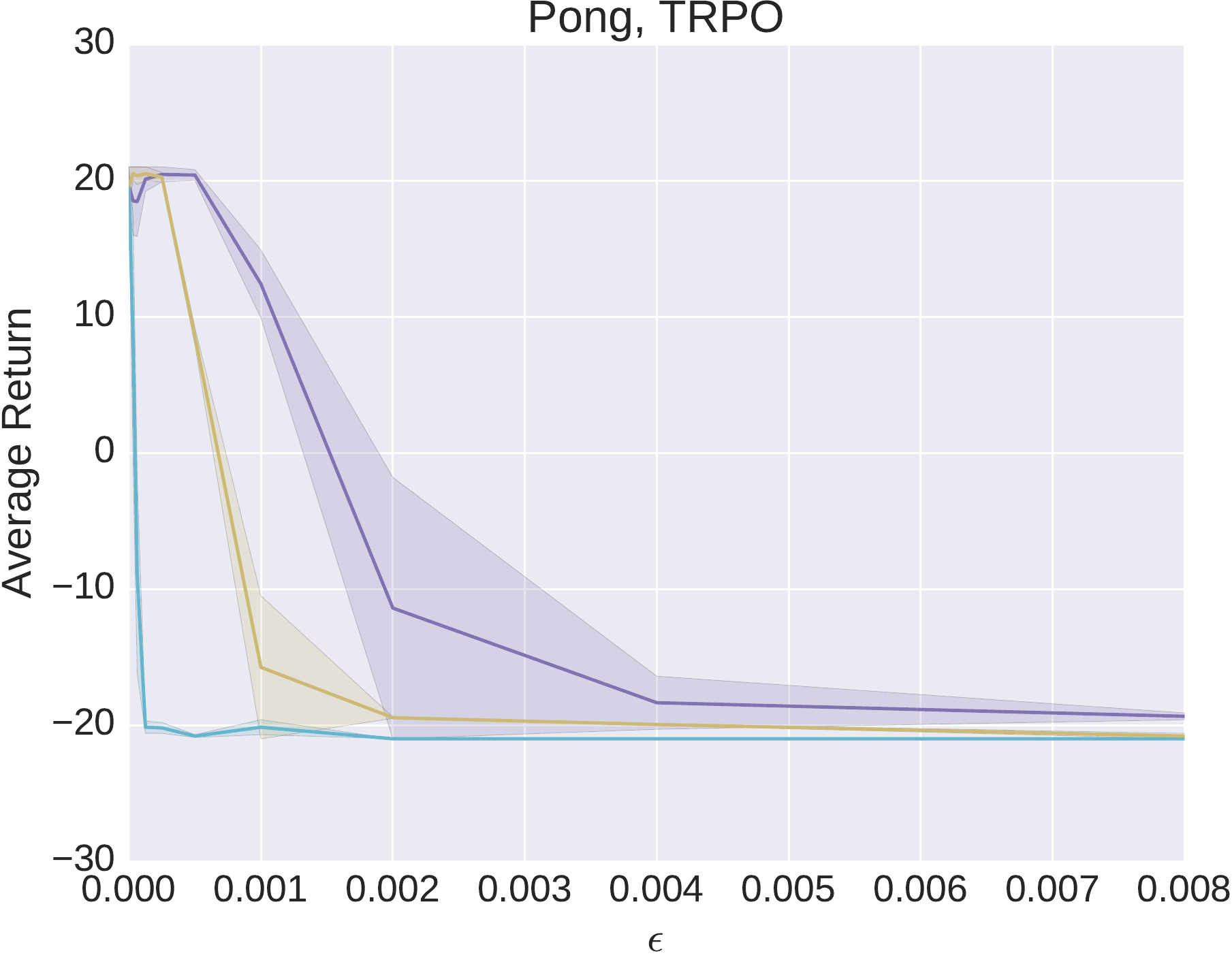}} &
\subfloat{\includegraphics[width = 1.33in,natwidth=529,natheight=406]{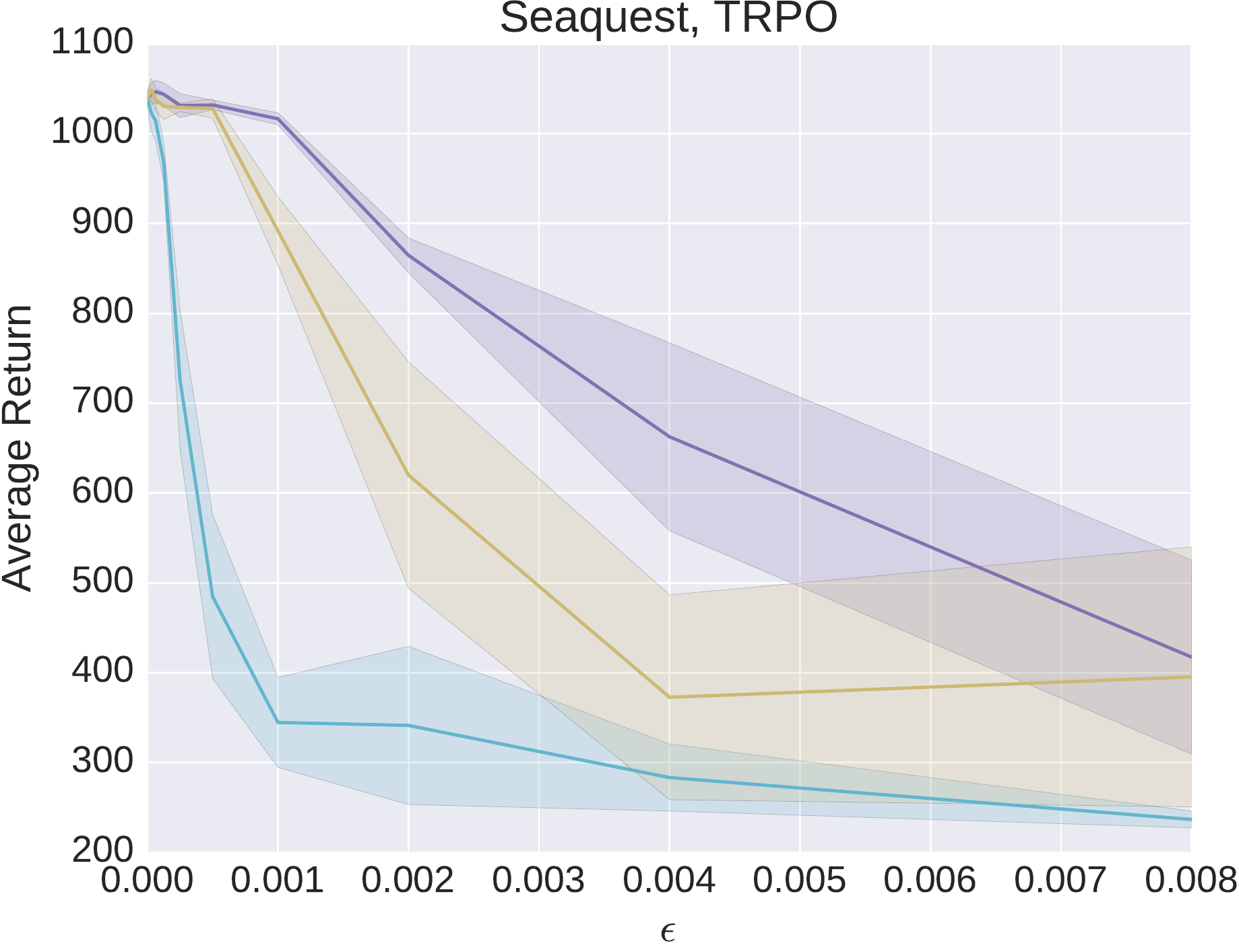}} &
\subfloat{\includegraphics[width = 1.33in,natwidth=529,natheight=406]{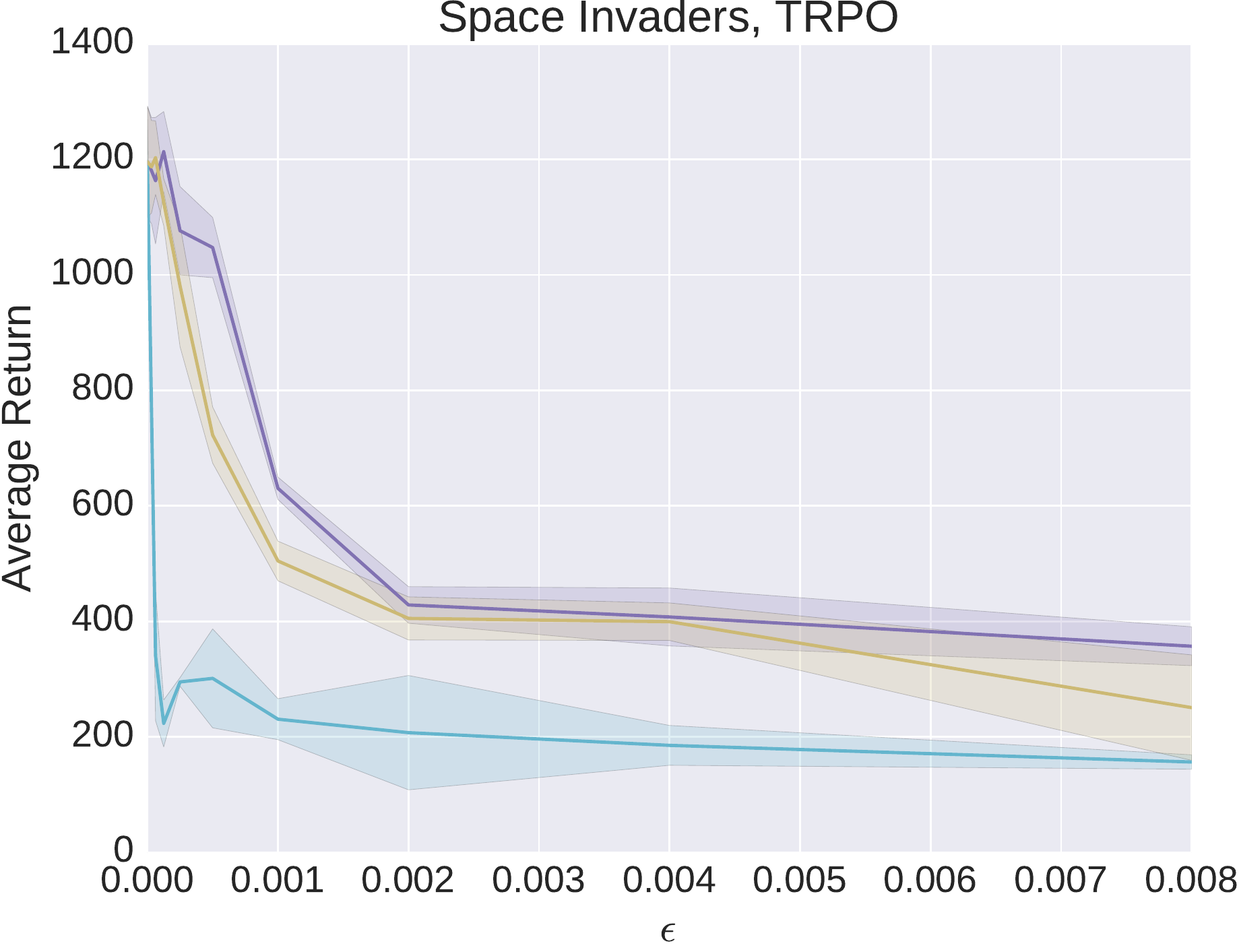}} \\
\addlinespace[-2ex]
\subfloat{\includegraphics[width = 1.33in,natwidth=529,natheight=406]{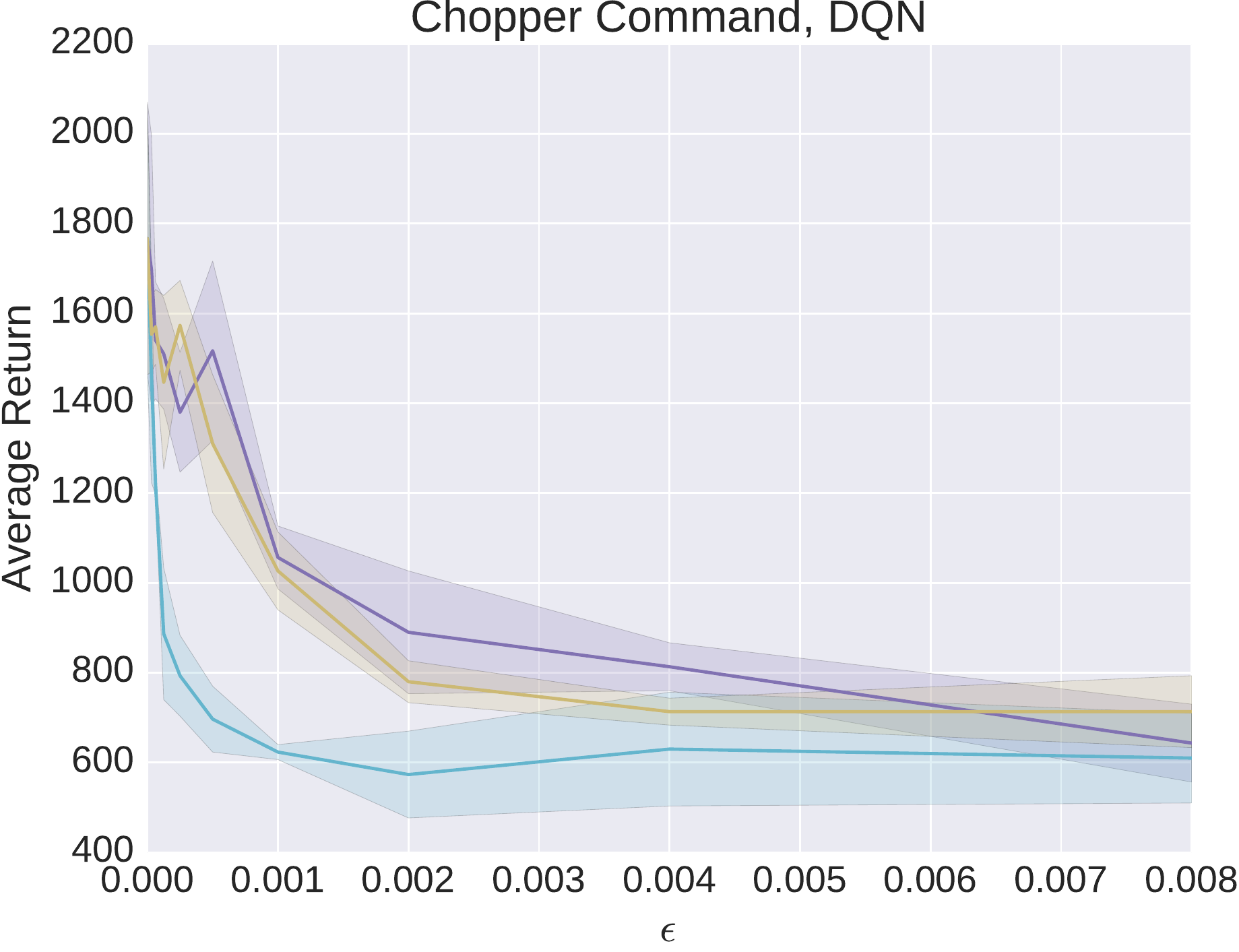}} &
\subfloat{\includegraphics[width = 1.33in,natwidth=520,natheight=406]{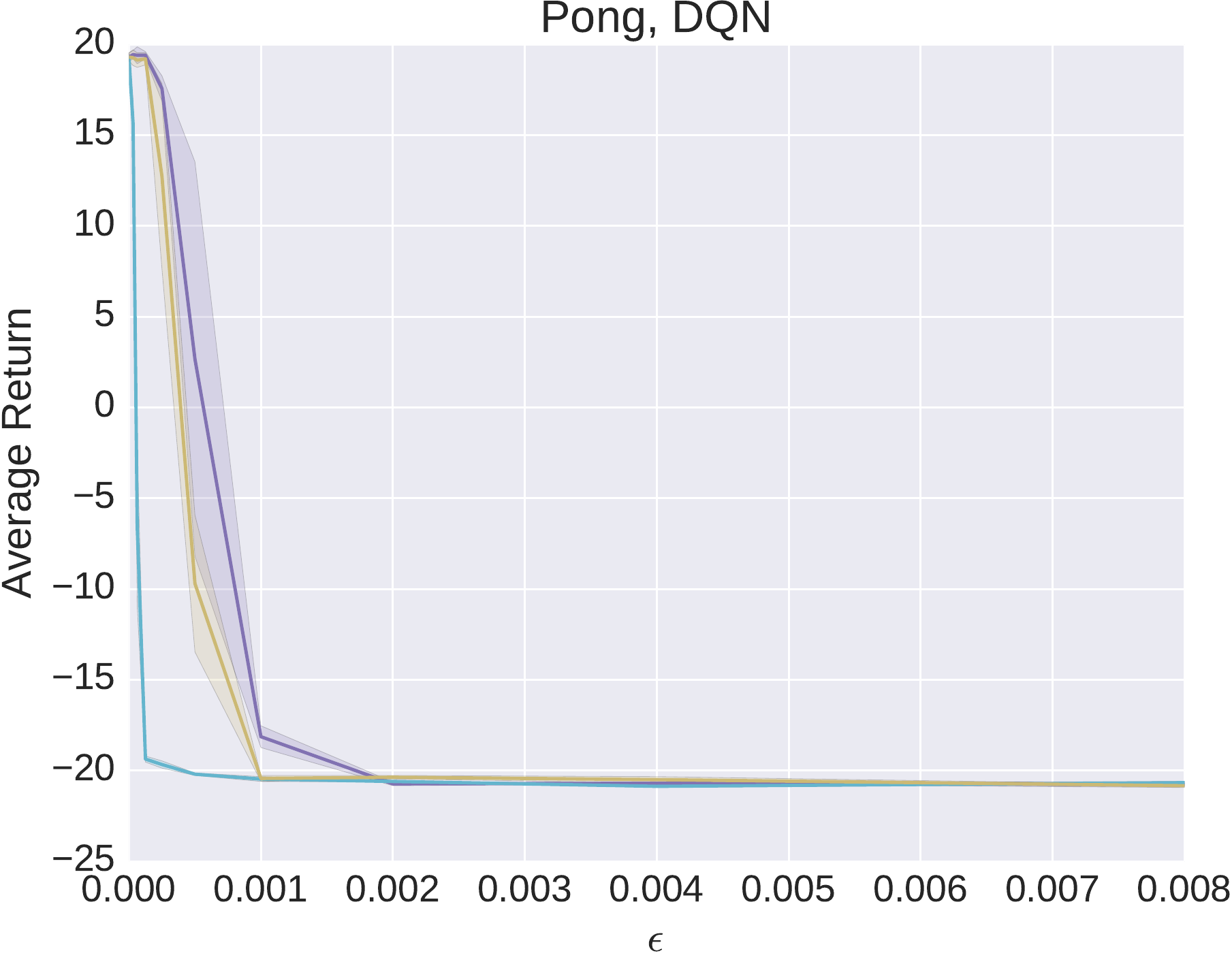}} &
\subfloat{\includegraphics[width = 1.33in,natwidth=529,natheight=406]{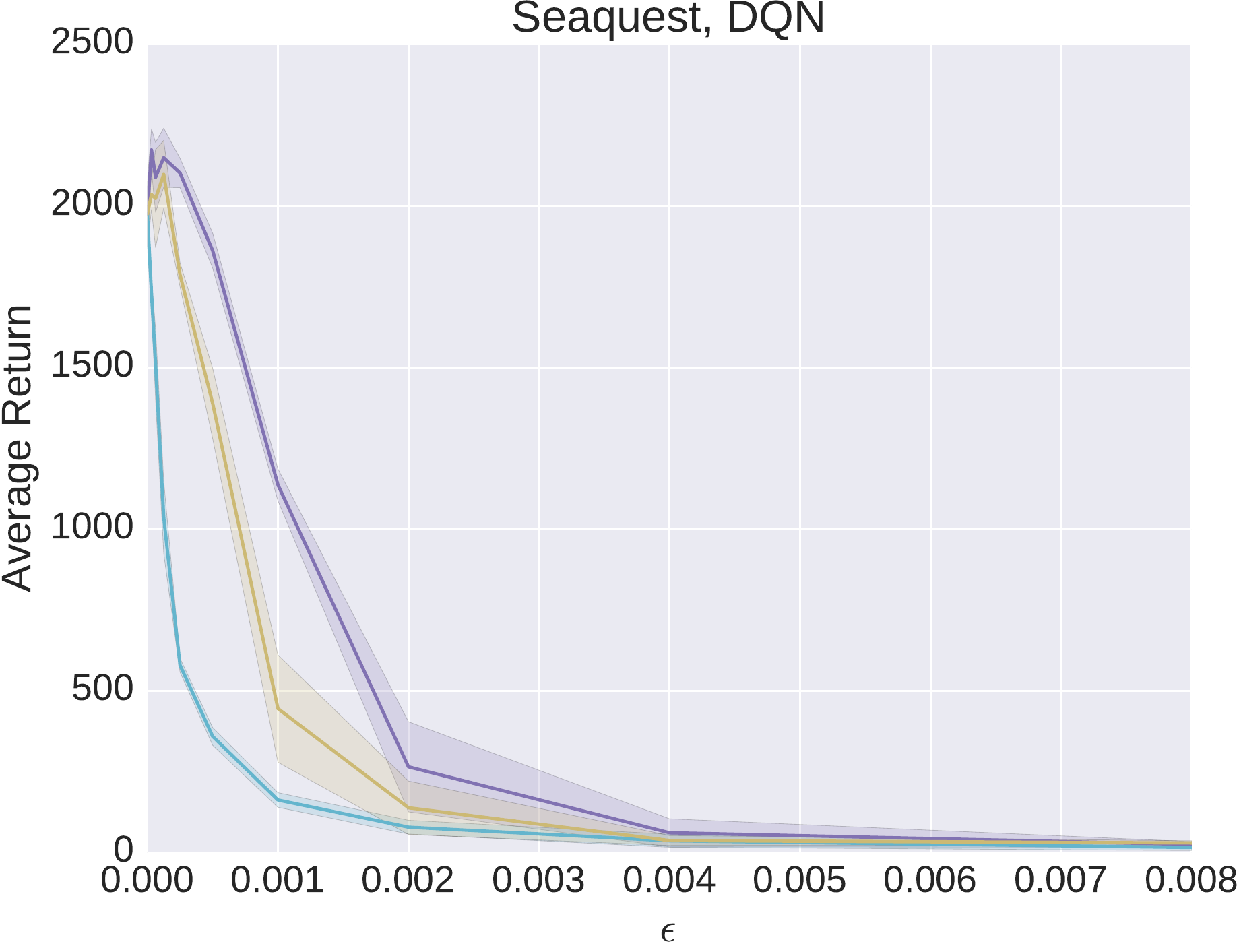}} &
\subfloat{\includegraphics[width = 1.33in,natwidth=520,natheight=406]{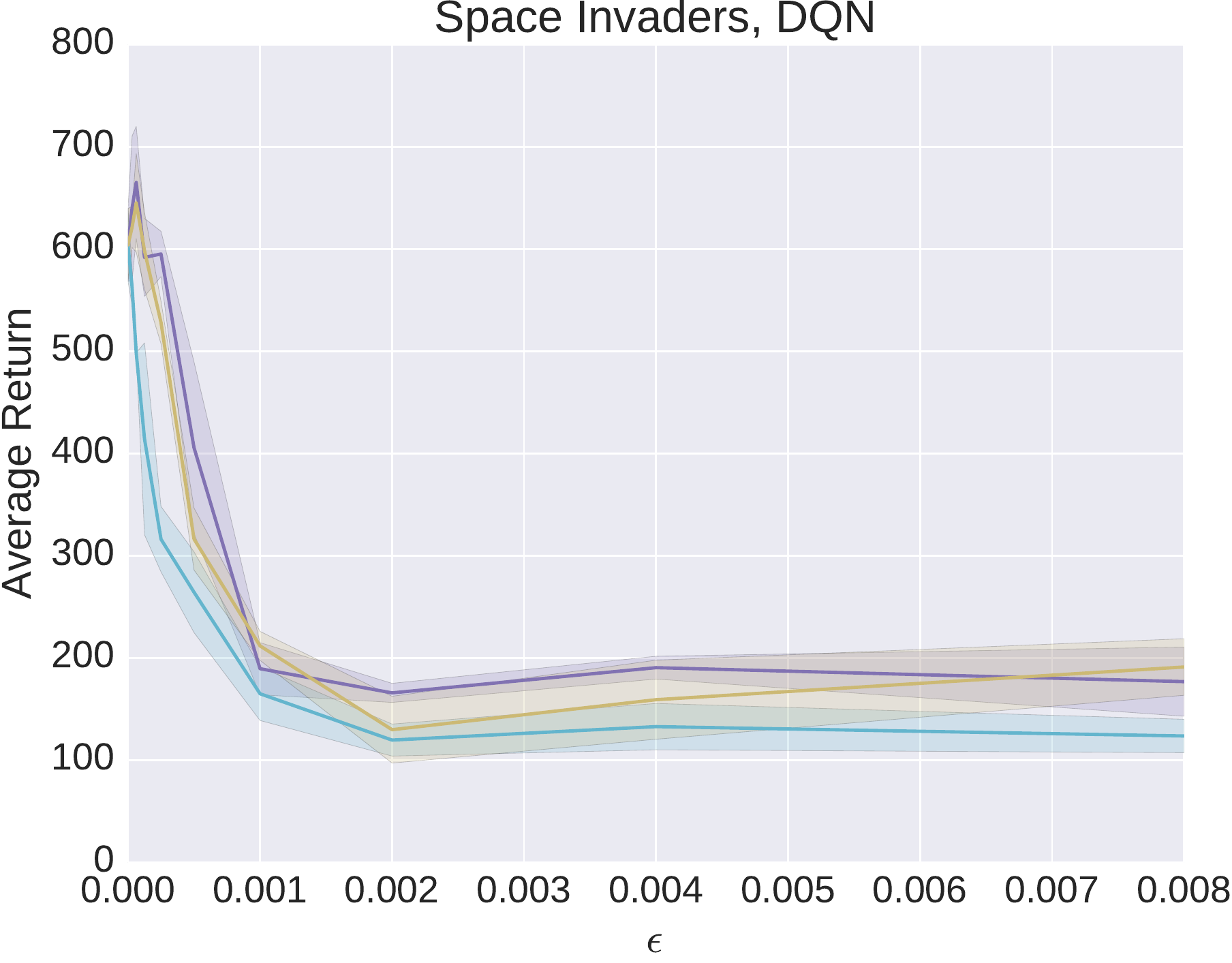}}
\end{tabular}
\caption{Comparison of the effectiveness of $\ell_\infty$, $\ell_2$, and $\ell_1$ FGSM adversaries on four Atari games trained with three learning algorithms. The average return is taken across ten trajectories. Constraint on FGSM perturbation: 
\crule[plot-linf]{0.3cm}{0.3cm} $\ell_\infty$-norm
\crule[plot-l2]{0.3cm}{0.3cm} $\ell_2$-norm
\crule[plot-l1]{0.3cm}{0.3cm} $\ell_1$-norm}
\label{fig:notransfer}
\end{figure*}

We find that regardless of which game the policy is trained for or how it is trained, it is indeed possible to significantly decrease the policy's performance through introducing relatively small perturbations in the inputs (\figref{fig:notransfer}). 

Notably, in many cases an $\ell_\infty$-norm FGSM adversary with $\epsilon = 0.001$ decreases the agent's performance by $50\%$ or more; when converted to 8-bit image encodings, these adversarial inputs are indistinguishable from the original inputs.

In cases where it is not essential for changes to be imperceptible, using an $\ell_1$-norm adversary may be a better choice: given the same $\epsilon$, $\ell_1$-norm adversaries are able to achieve the most significant decreases in agent performance. They are able to sharply decrease the agent's performance just by changing a few pixels (by large amounts). 

We see that policies trained with A3C, TRPO, and DQN are all susceptible to adversarial inputs. Interestingly, policies trained with DQN are more susceptible, especially to $\ell_\infty$-norm FGSM perturbations on Pong, Seaquest, and Space Invaders.

\setlength\tabcolsep{1.5pt}
\setlength\tabcolsep{6pt} 

\subsection{Vulnerability to Black-Box Attacks}
In practice, it is often the case that an adversary does not have complete access to the neural network of the target policy~\cite{papernot2016practical}. This threat model is frequently referred to as  
a black-box scenario. We investigate how vulnerable neural network policies are to black-box attacks of the following two variants:
\begin{enumerate}
    \item The adversary has access to the training environment and knowledge of the training algorithm and hyperparameters. It knows the neural network architecture of the target policy network, but not its random initialization. We will refer to this as transferability across policies.
    \item The adversary additionally has no knowledge of the training algorithm or hyperparameters. We will refer to this as transferability across algorithms.
\end{enumerate}

\subsubsection{Transferability Across Policies}
To explore transferability of adversarial examples across policies, we generate adversarial perturbations for the target policy using one of the other top-performing policies trained with the same algorithm for the same task. We test all adversary-target combinations of top-performing policies trained with the same algorithm, for each combination of task, learning algorithm, and type of adversary.

\subsubsection{Transferability Across Training Algorithms}
To explore transferability of adversarial examples across training algorithms, we generate adversarial perturbations for the target policy using one of the top-performing policies trained with a \emph{different} algorithm. Similarly, we test all adversary-target combinations of top-performing policies trained with different algorithms, for each combination of task and type of adversary.

\subsubsection{Observations}
\begin{figure*}[t!]
\centering
\begin{tabular}{cccc}
\subfloat{\includegraphics[width = 1.33in,natwidth=521,natheight=402]{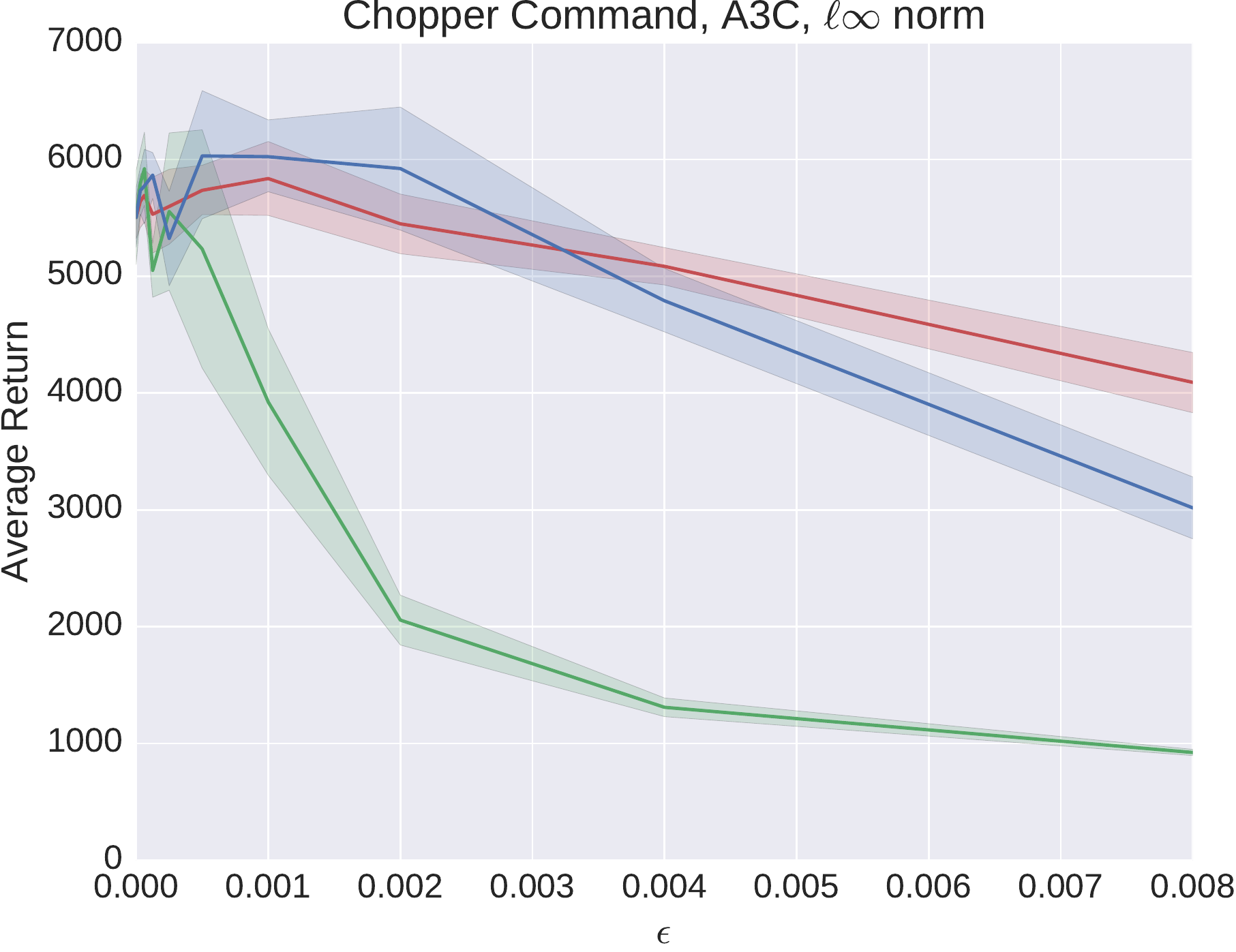}} &
\subfloat{\includegraphics[width = 1.33in,natwidth=514,natheight=402]{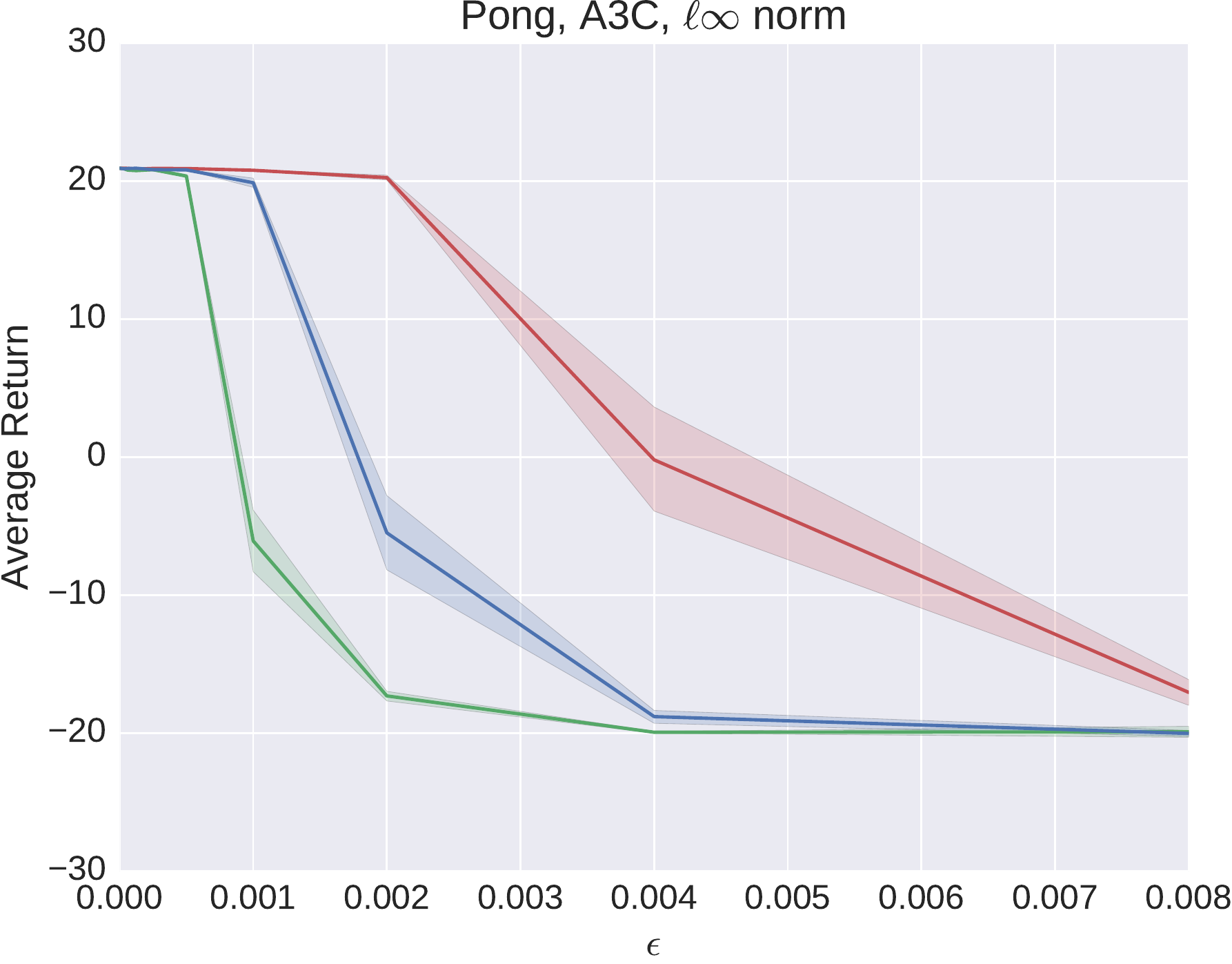}} &
\subfloat{\includegraphics[width = 1.33in,natwidth=521,natheight=402]{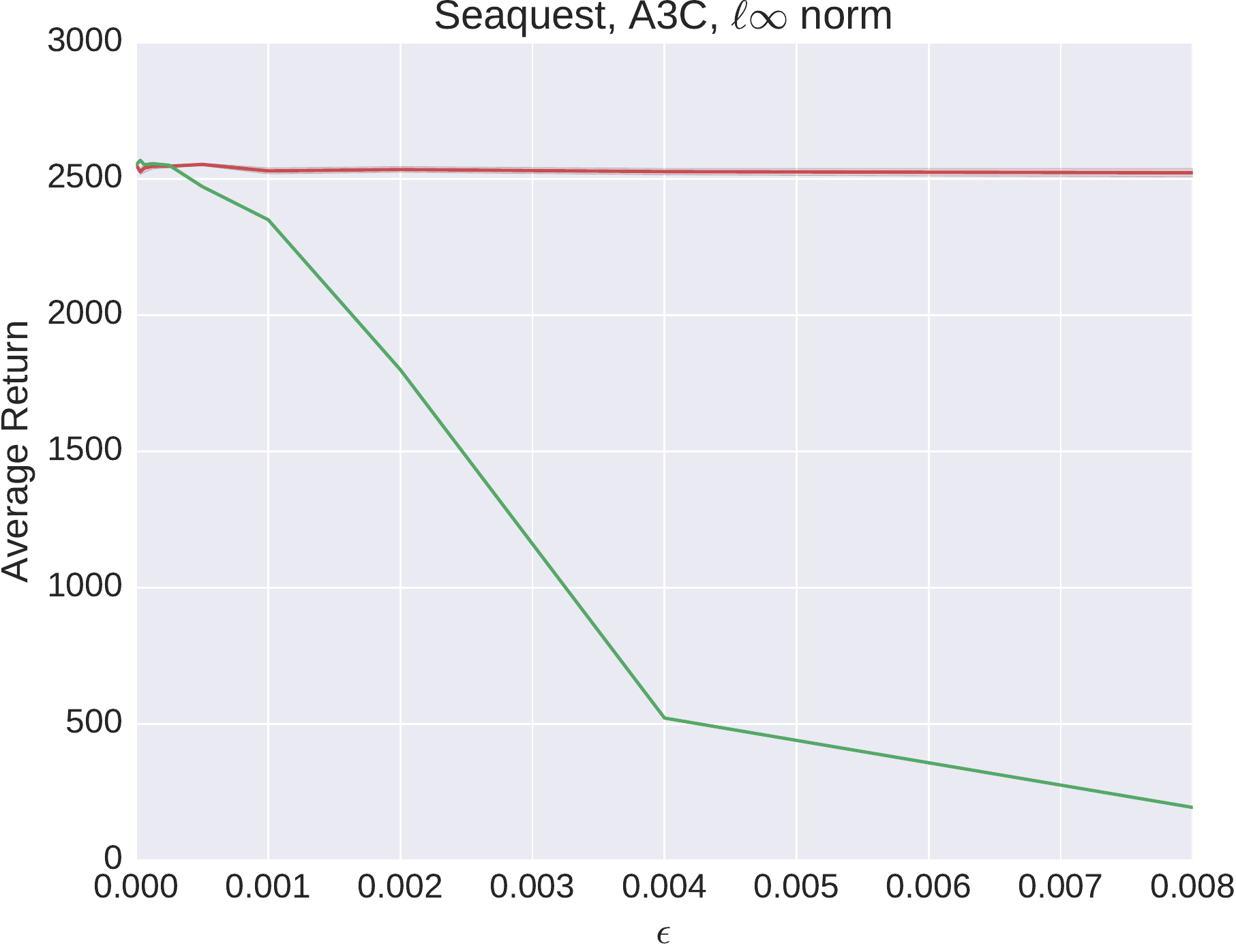}} &
\subfloat{\includegraphics[width = 1.33in,natwidth=521,natheight=402]{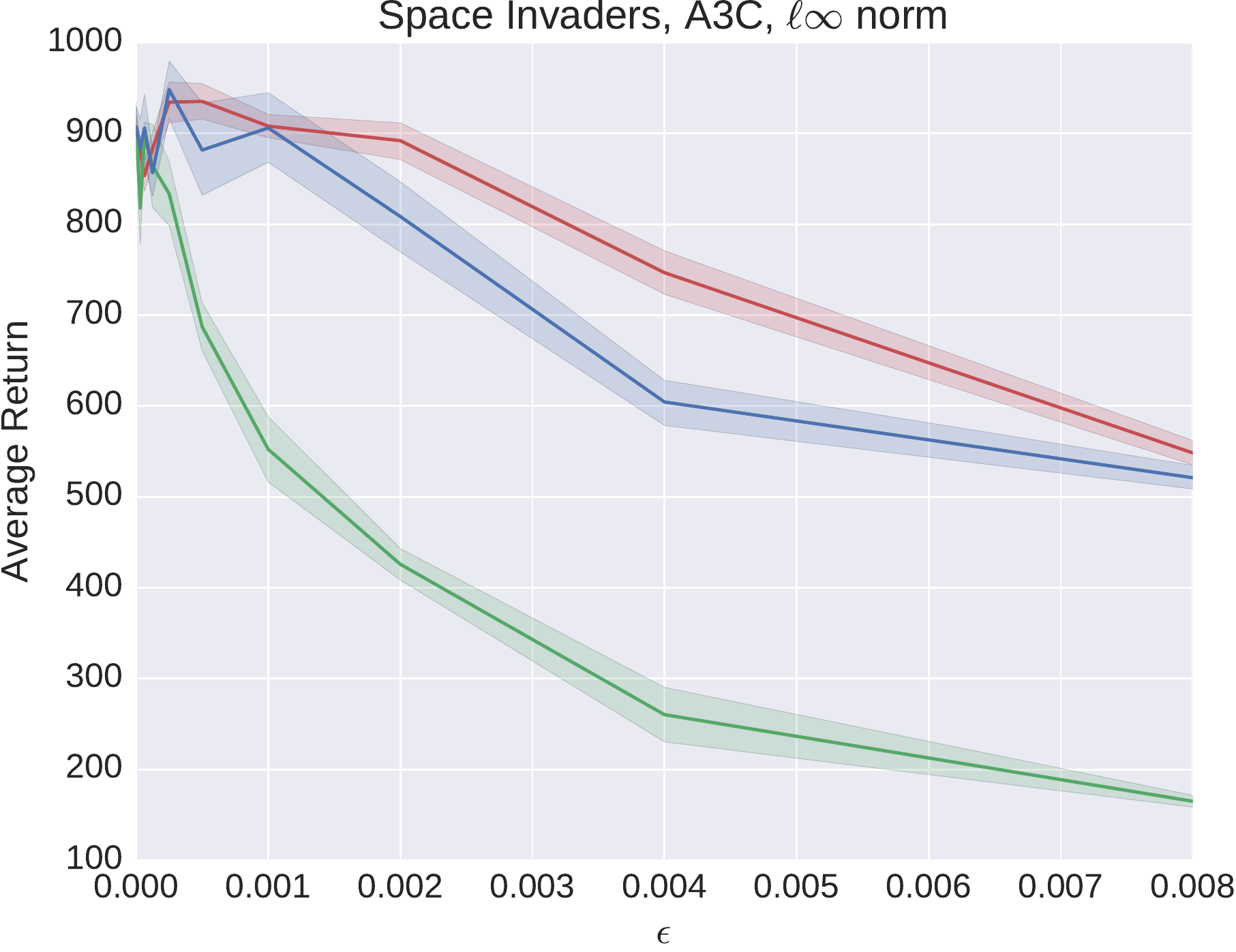}} \\
\addlinespace[-2ex]
\subfloat{\includegraphics[width = 1.33in,natwidth=521,natheight=402]{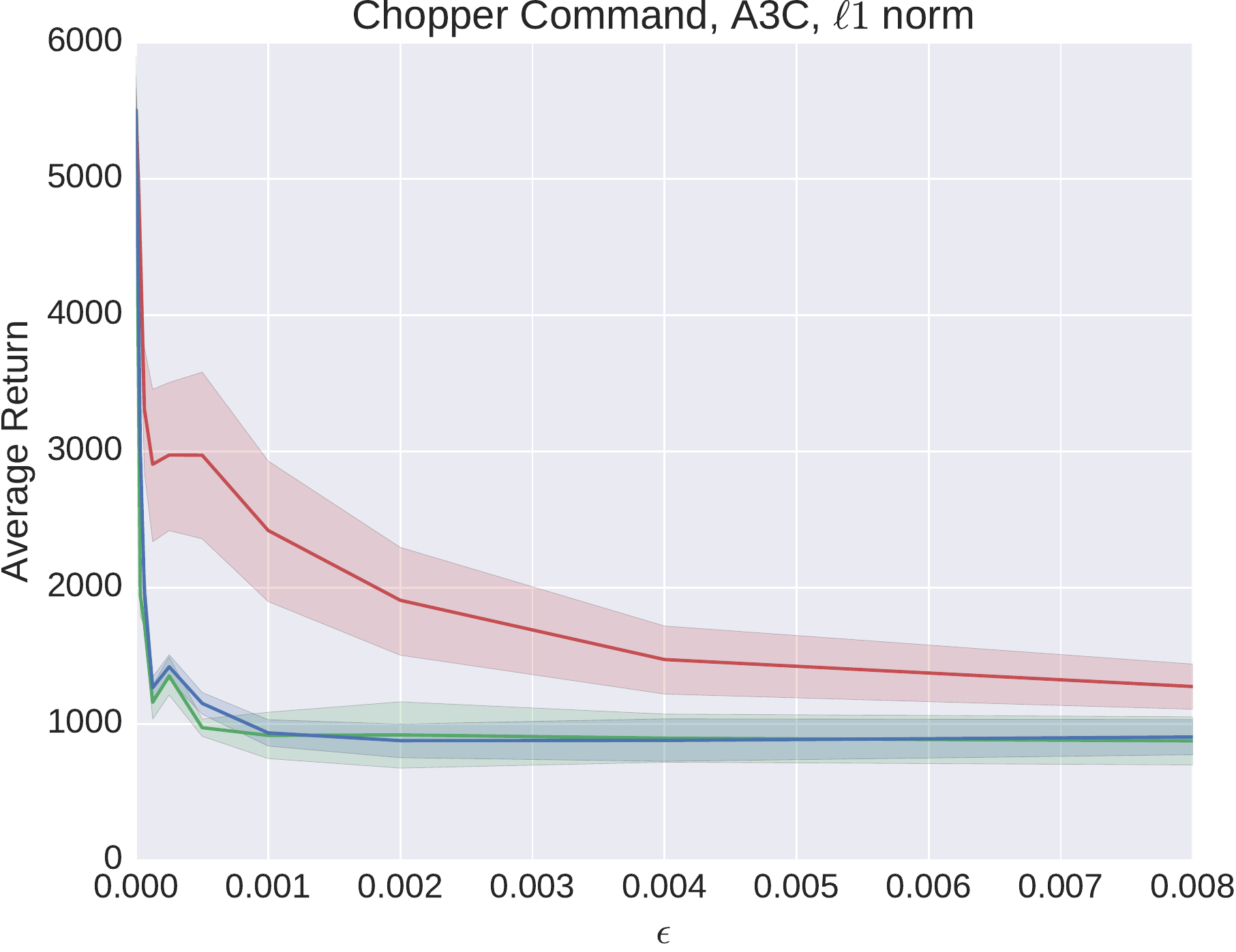}} &
\subfloat{\includegraphics[width = 1.33in,natwidth=514,natheight=402]{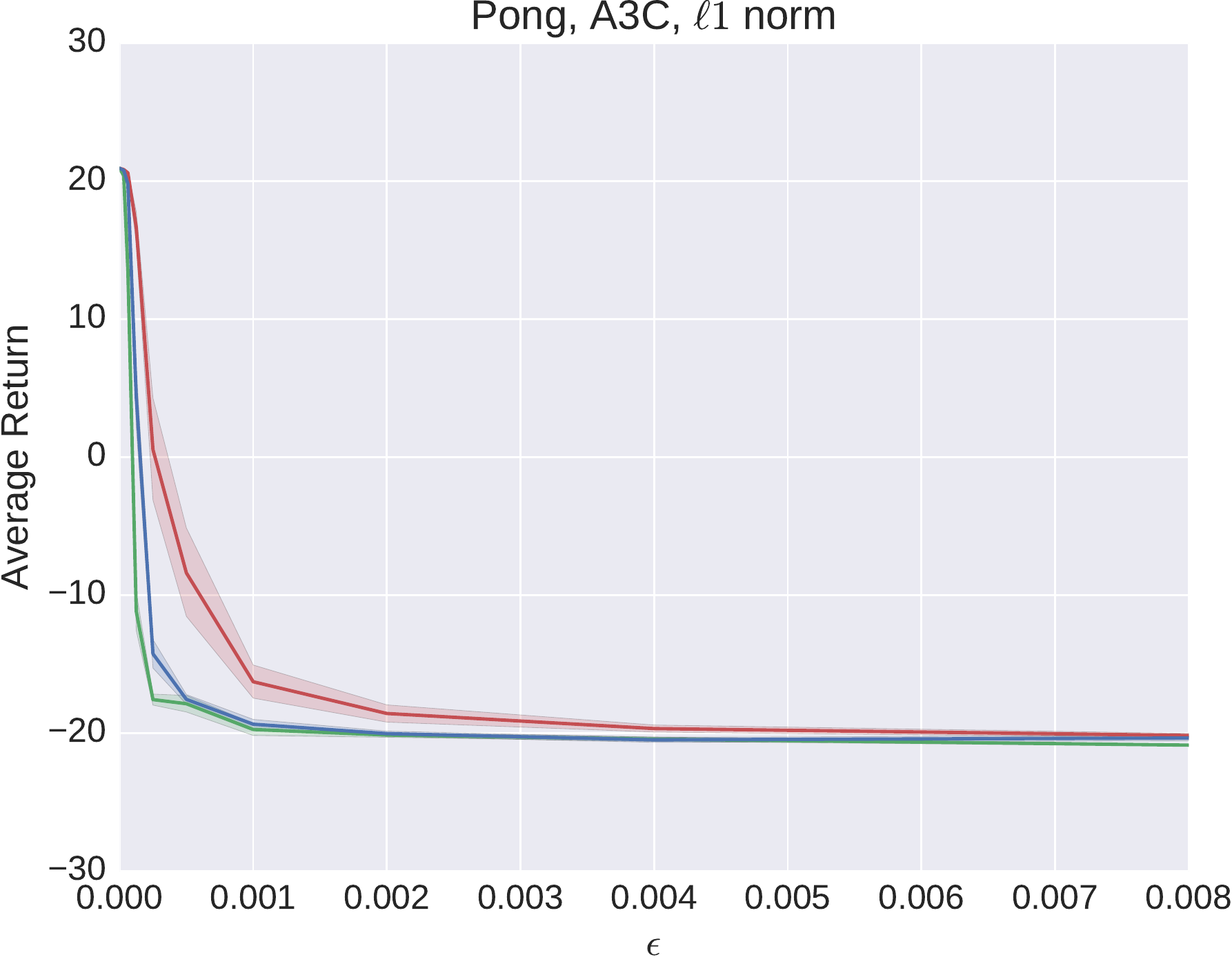}} &
\subfloat{\includegraphics[width = 1.33in,natwidth=521,natheight=402]{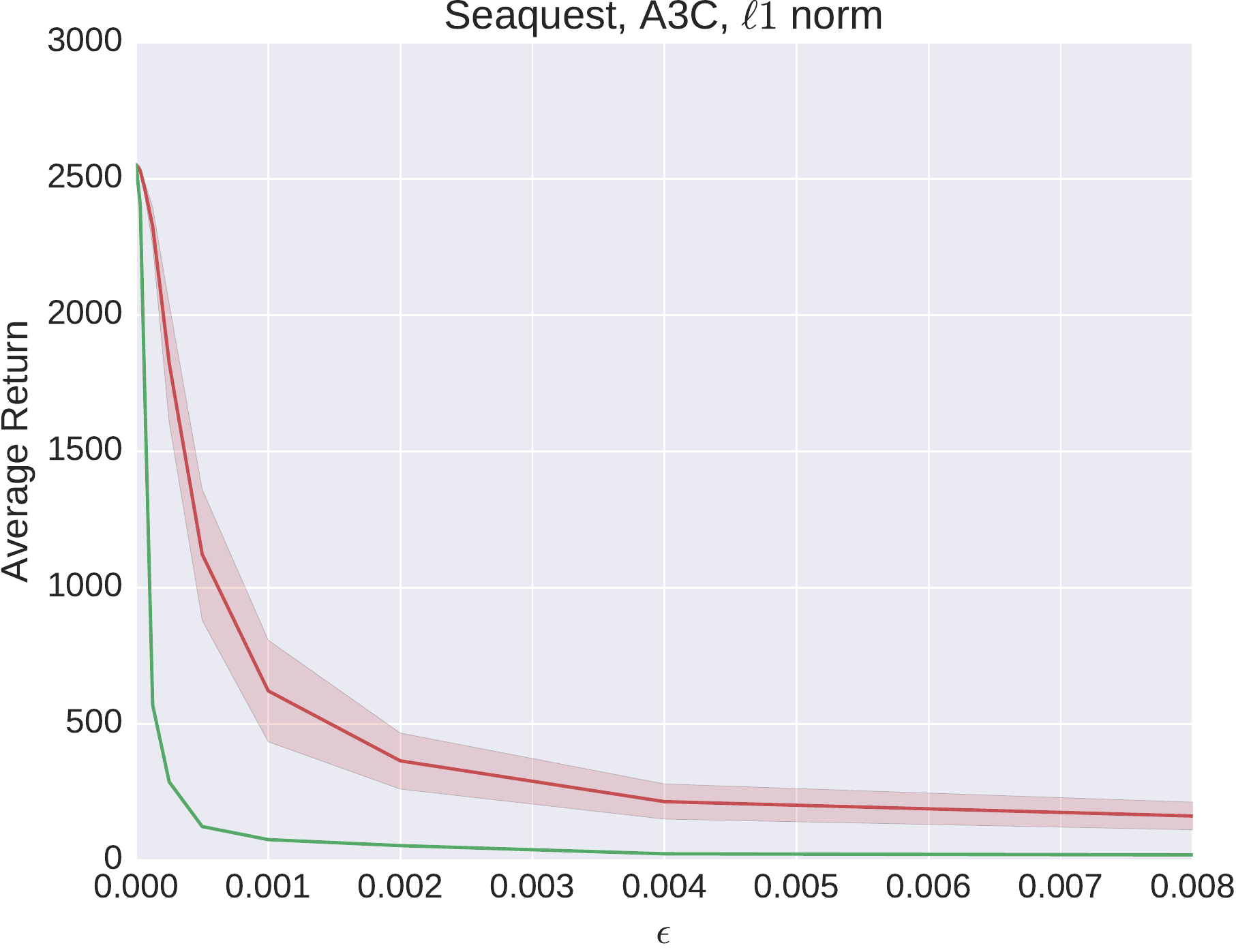}} &
\subfloat{\includegraphics[width = 1.33in,natwidth=521,natheight=402]{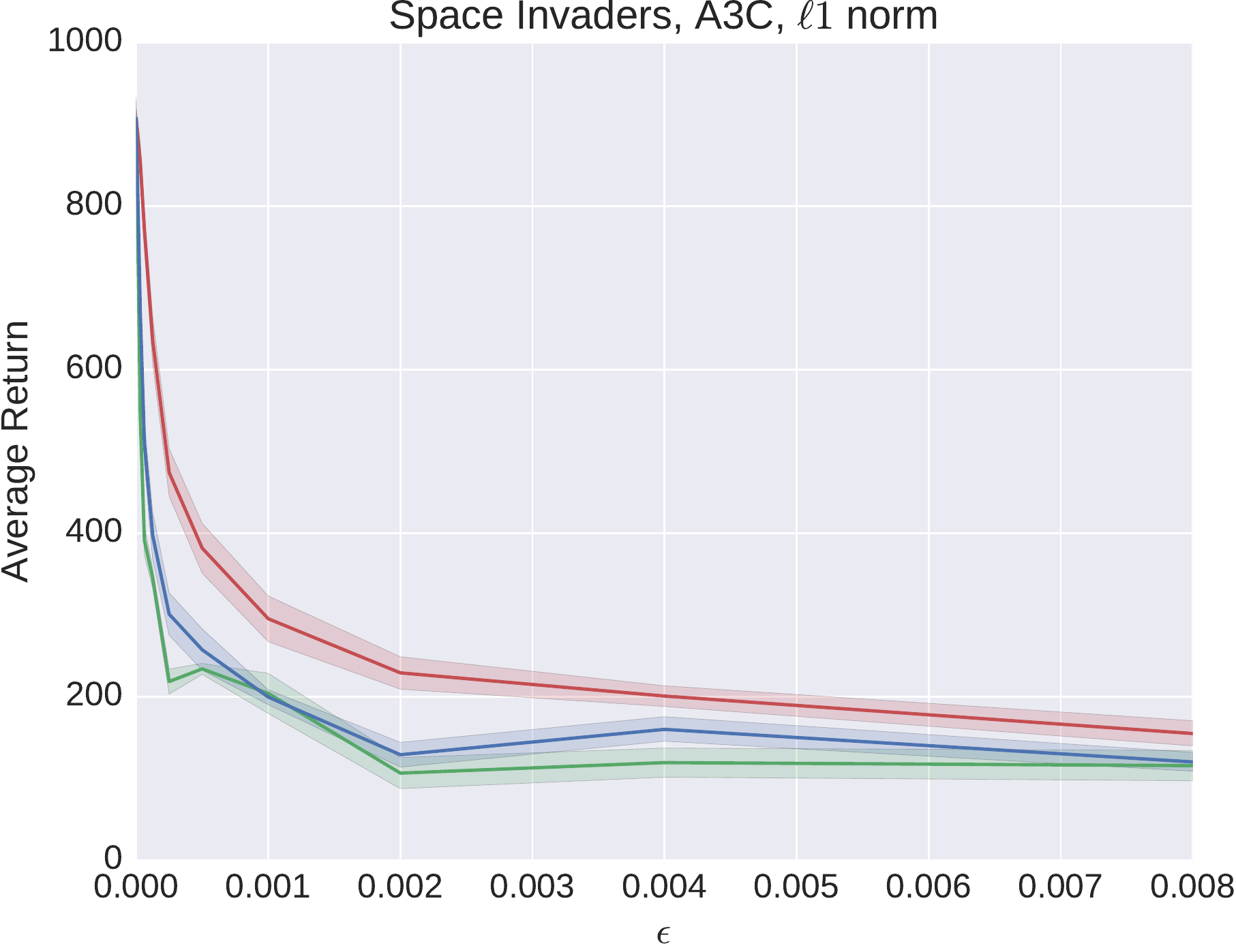}} \\
\addlinespace[-2ex]
\subfloat{\includegraphics[width = 1.33in,natwidth=521,natheight=402]{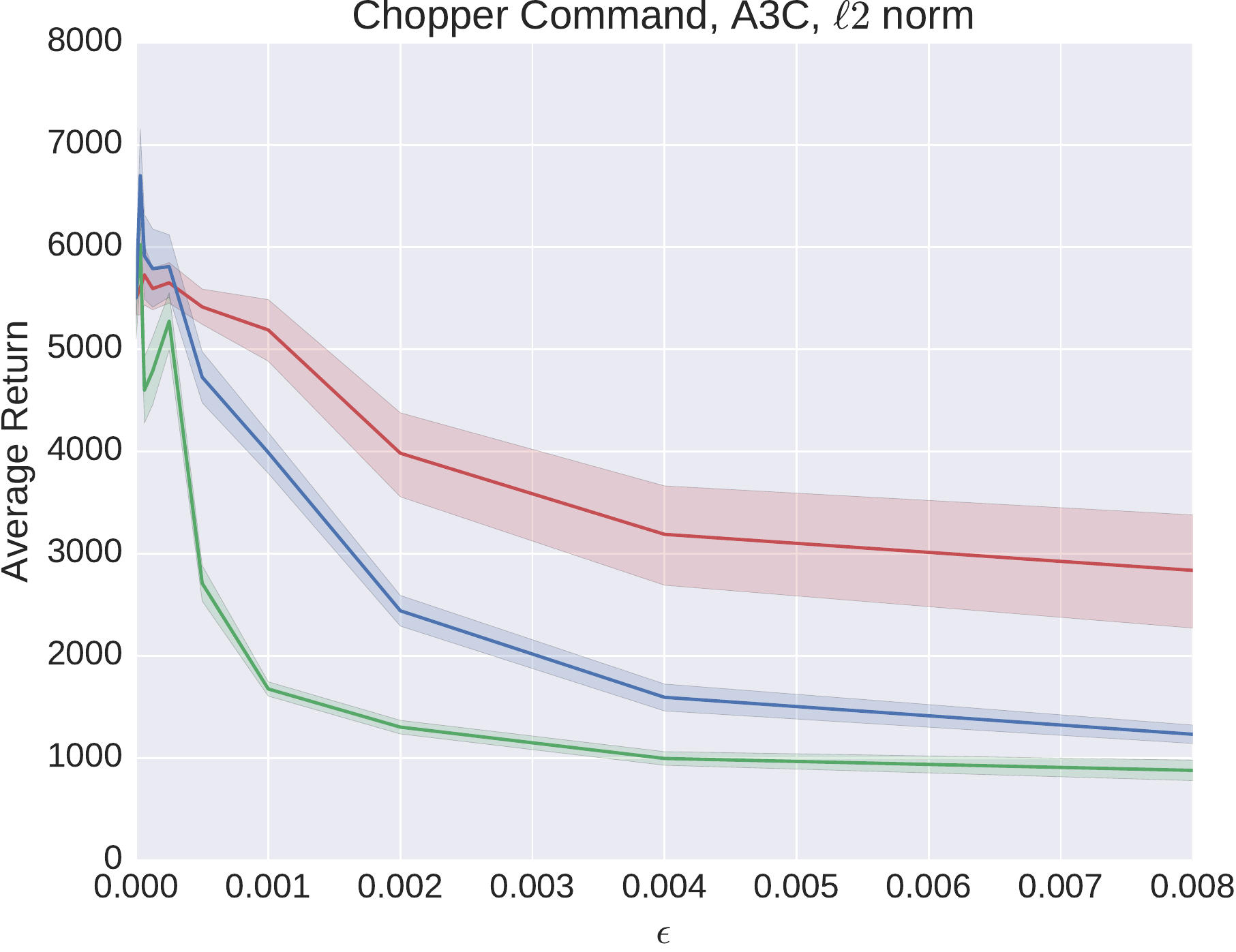}} &
\subfloat{\includegraphics[width = 1.33in,natwidth=514,natheight=402]{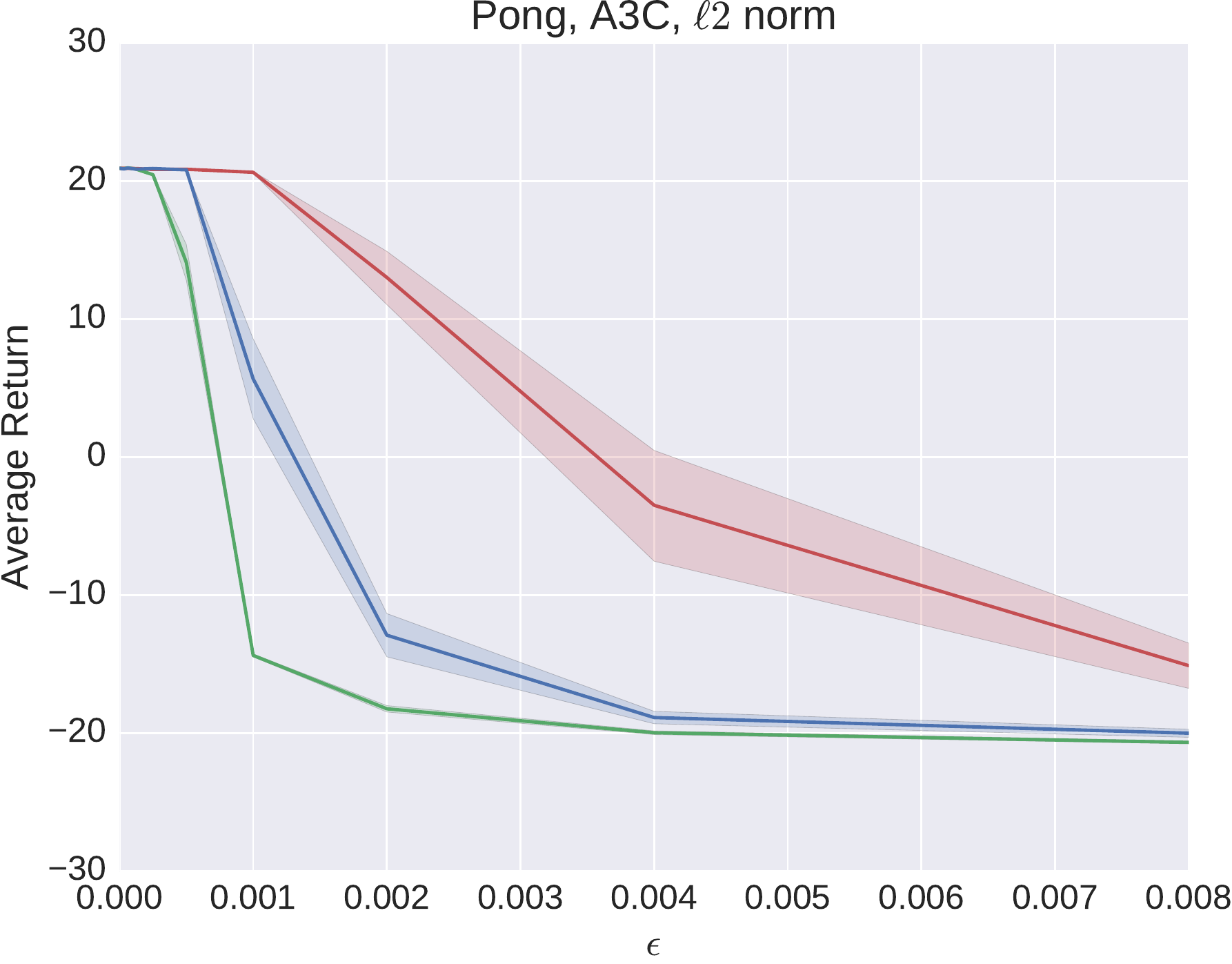}} &
\subfloat{\includegraphics[width = 1.33in,natwidth=521,natheight=402]{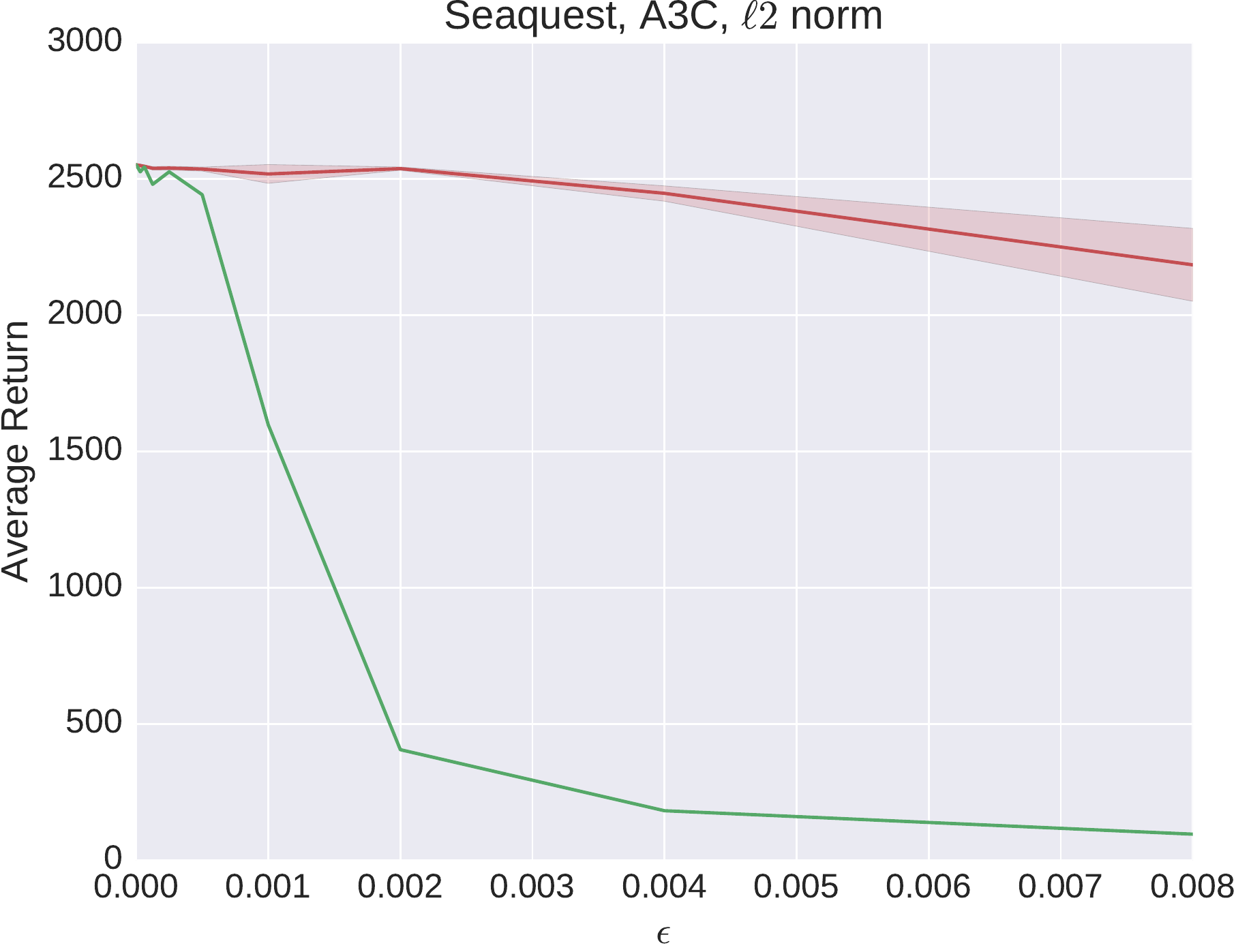}} &
\subfloat{\includegraphics[width = 1.33in,natwidth=521,natheight=402]{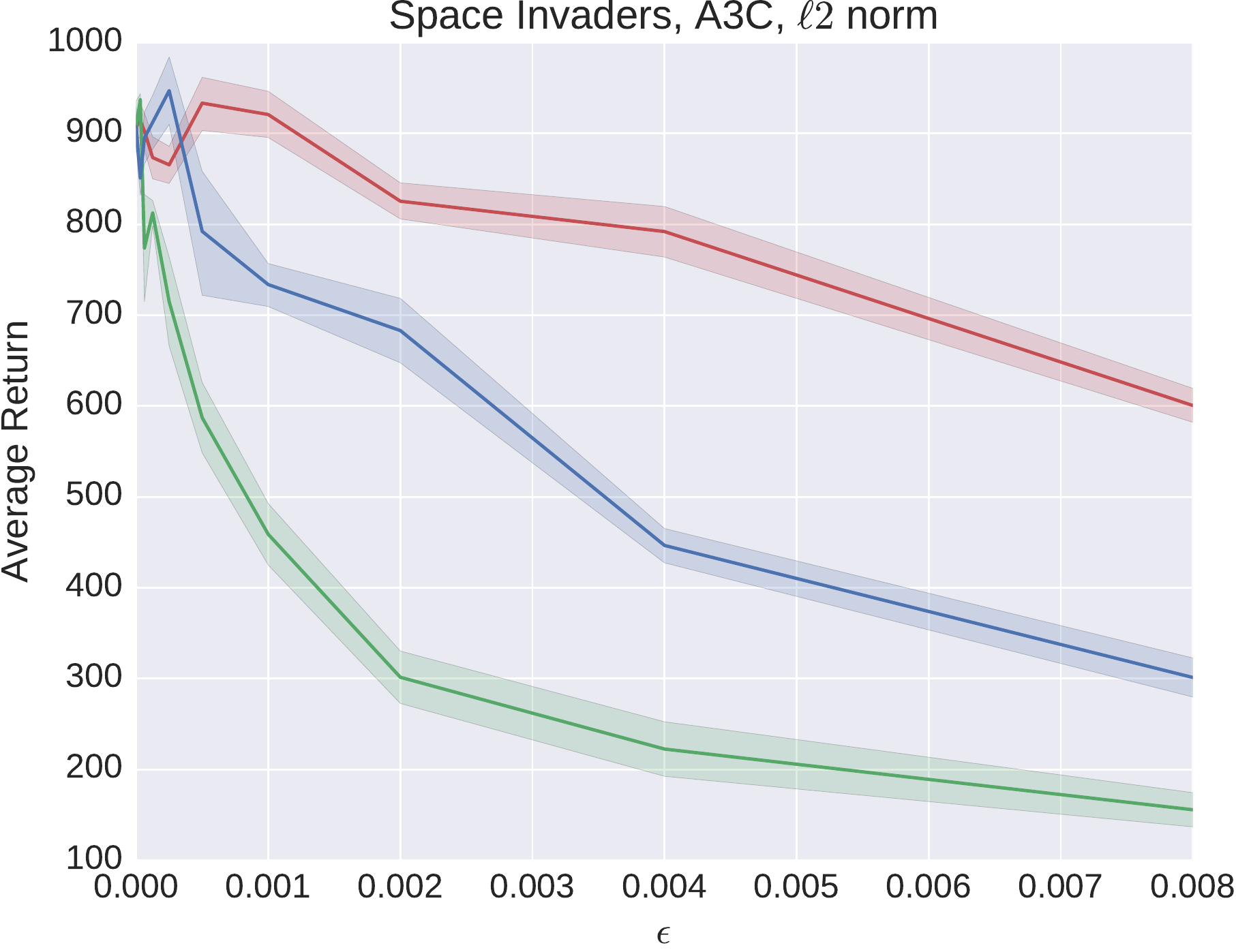}} 
\end{tabular}
\caption{Transferability of adversarial inputs for policies trained with A3C. Type of transfer:
\crule[plot-algo]{0.3cm}{0.3cm} algorithm
\crule[plot-policy]{0.3cm}{0.3cm} policy
\crule[plot-none]{0.3cm}{0.3cm} none}
\label{fig:transfer-A3C}
\end{figure*}

\begin{figure*}[t!]
\centering
\begin{tabular}{cccc}
\subfloat{\includegraphics[width = 1.33in,natwidth=521,natheight=402]{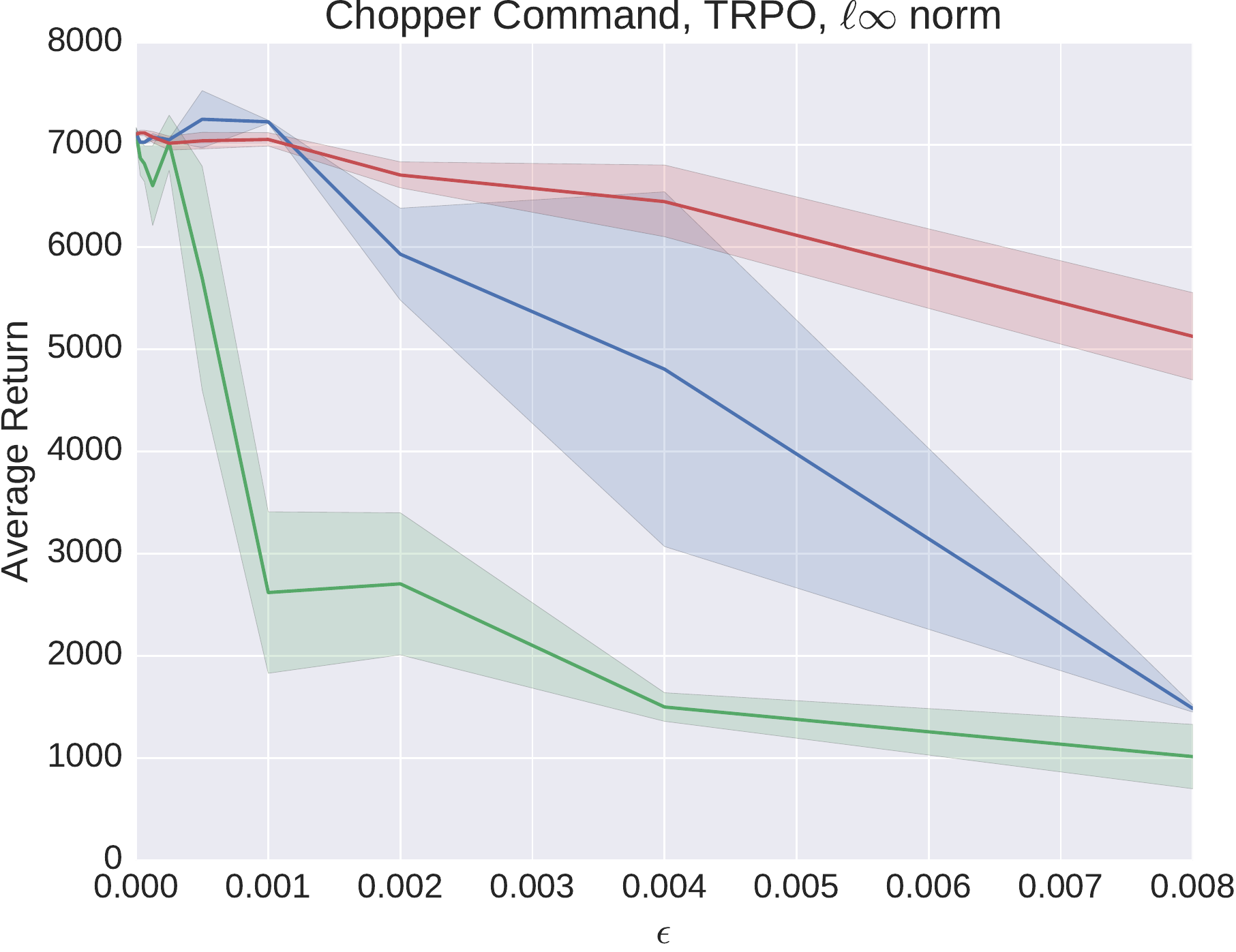}} &
\subfloat{\includegraphics[width = 1.33in,natwidth=514,natheight=402]{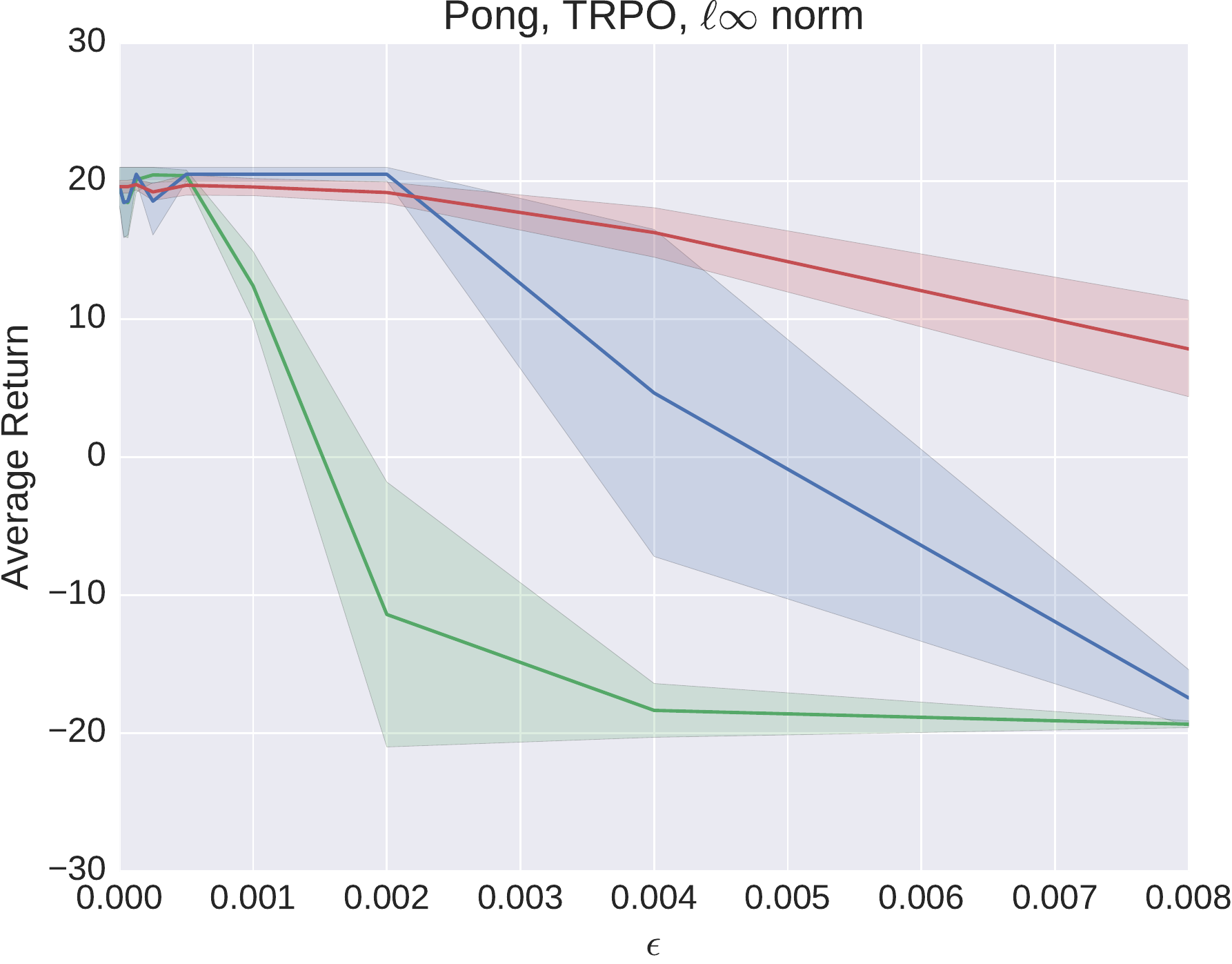}} &
\subfloat{\includegraphics[width = 1.33in,natwidth=521,natheight=402]{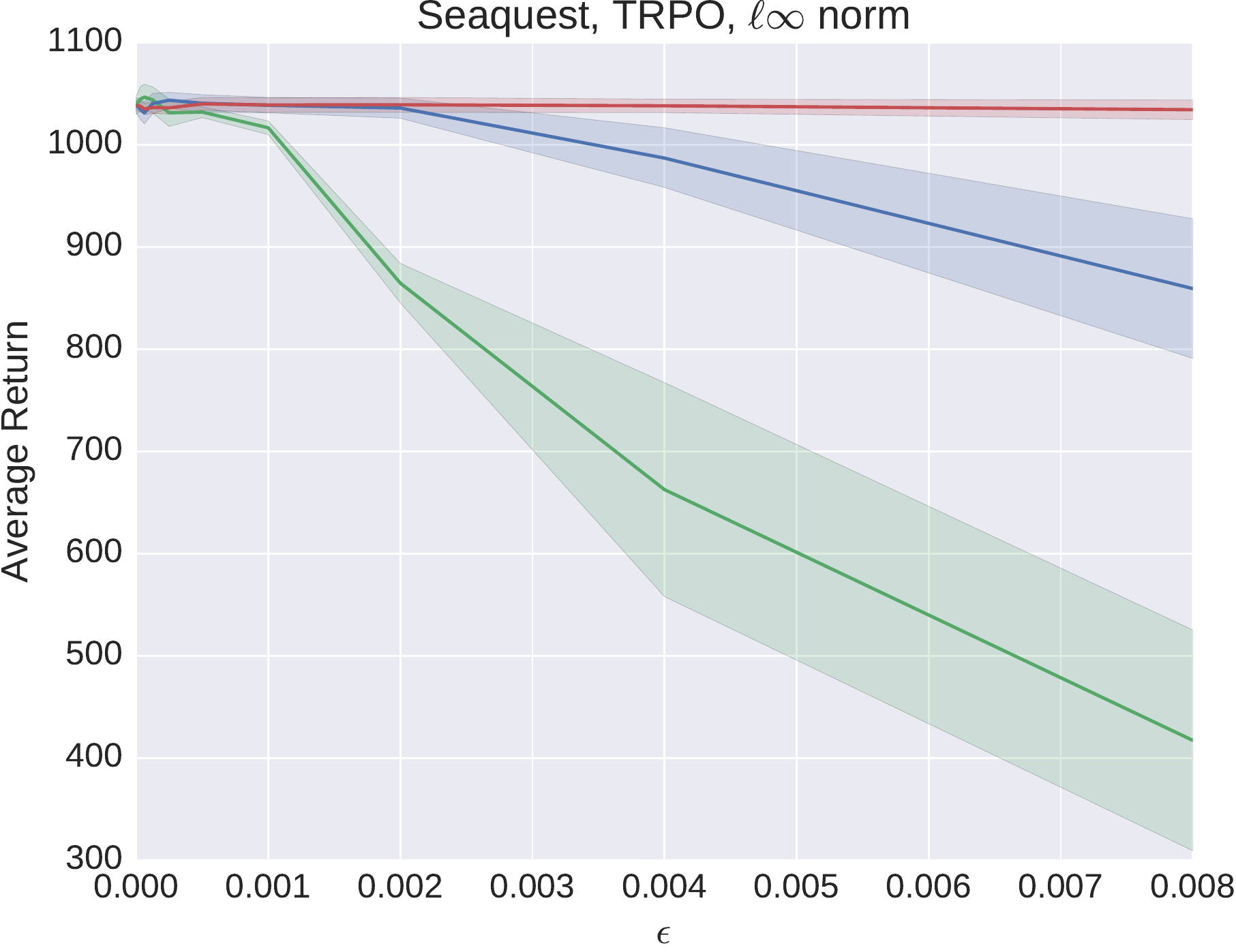}} &
\subfloat{\includegraphics[width = 1.33in,natwidth=521,natheight=402]{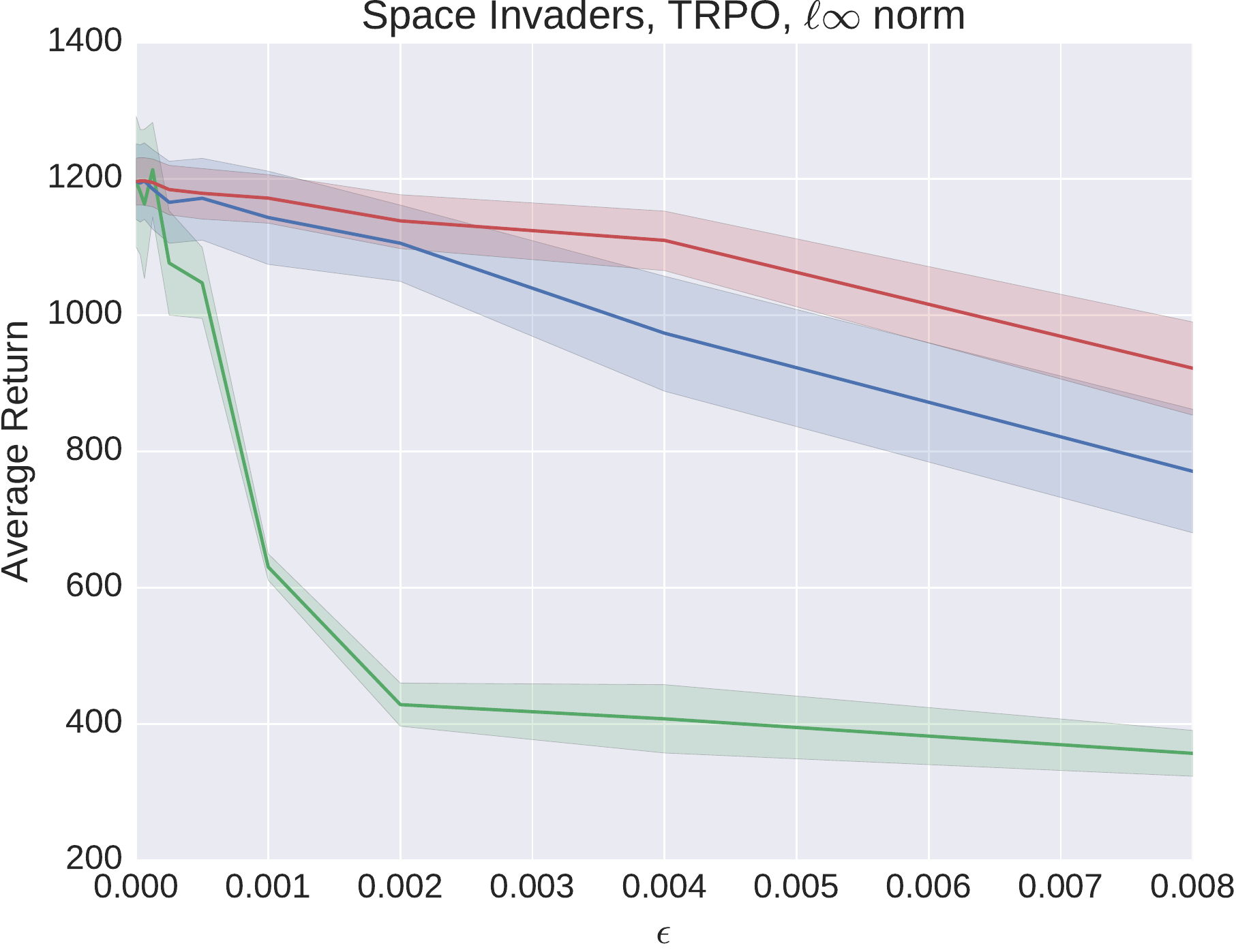}} \\
\addlinespace[-2ex]
\subfloat{\includegraphics[width = 1.33in,natwidth=521,natheight=402]{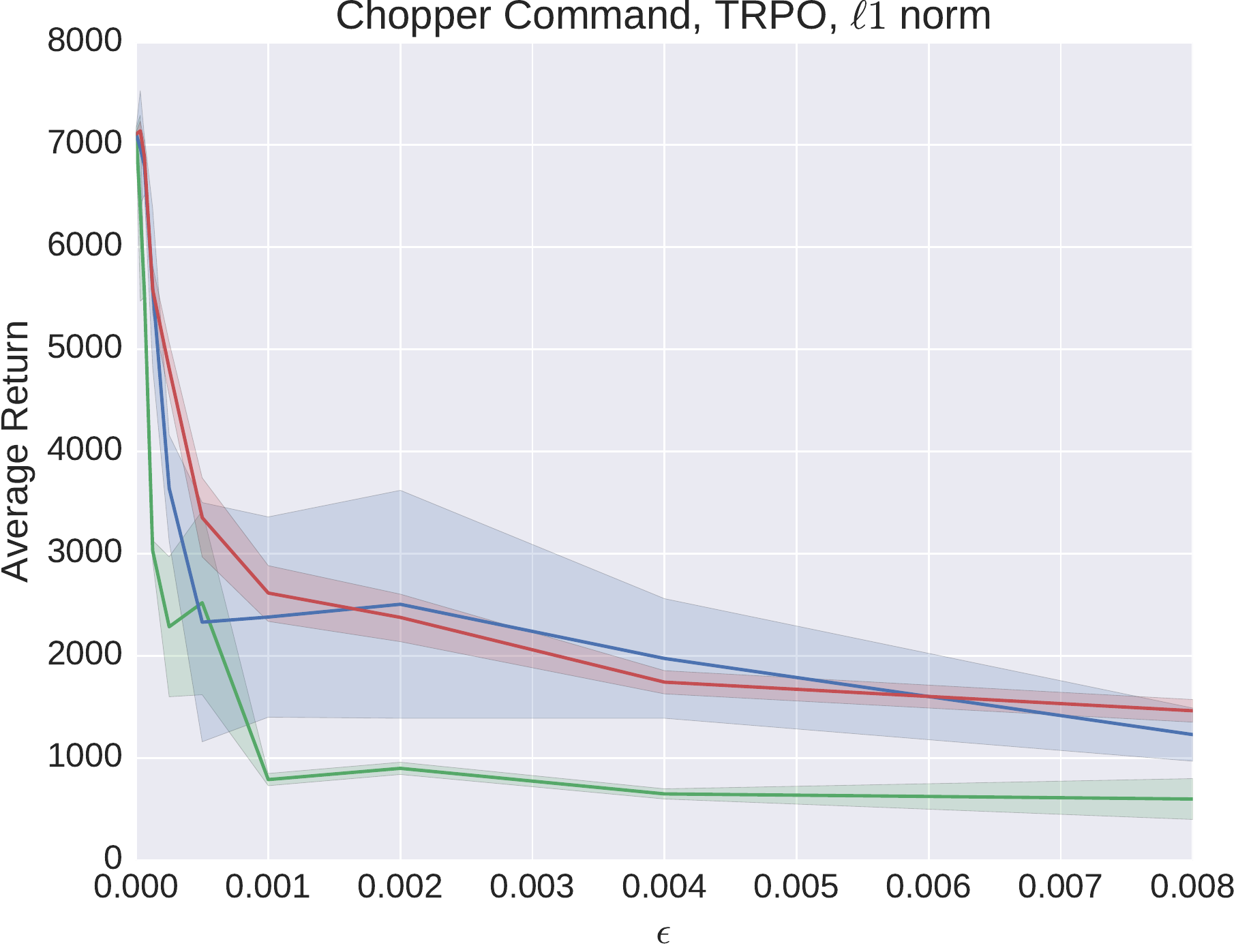}} &
\subfloat{\includegraphics[width = 1.33in,natwidth=514,natheight=402]{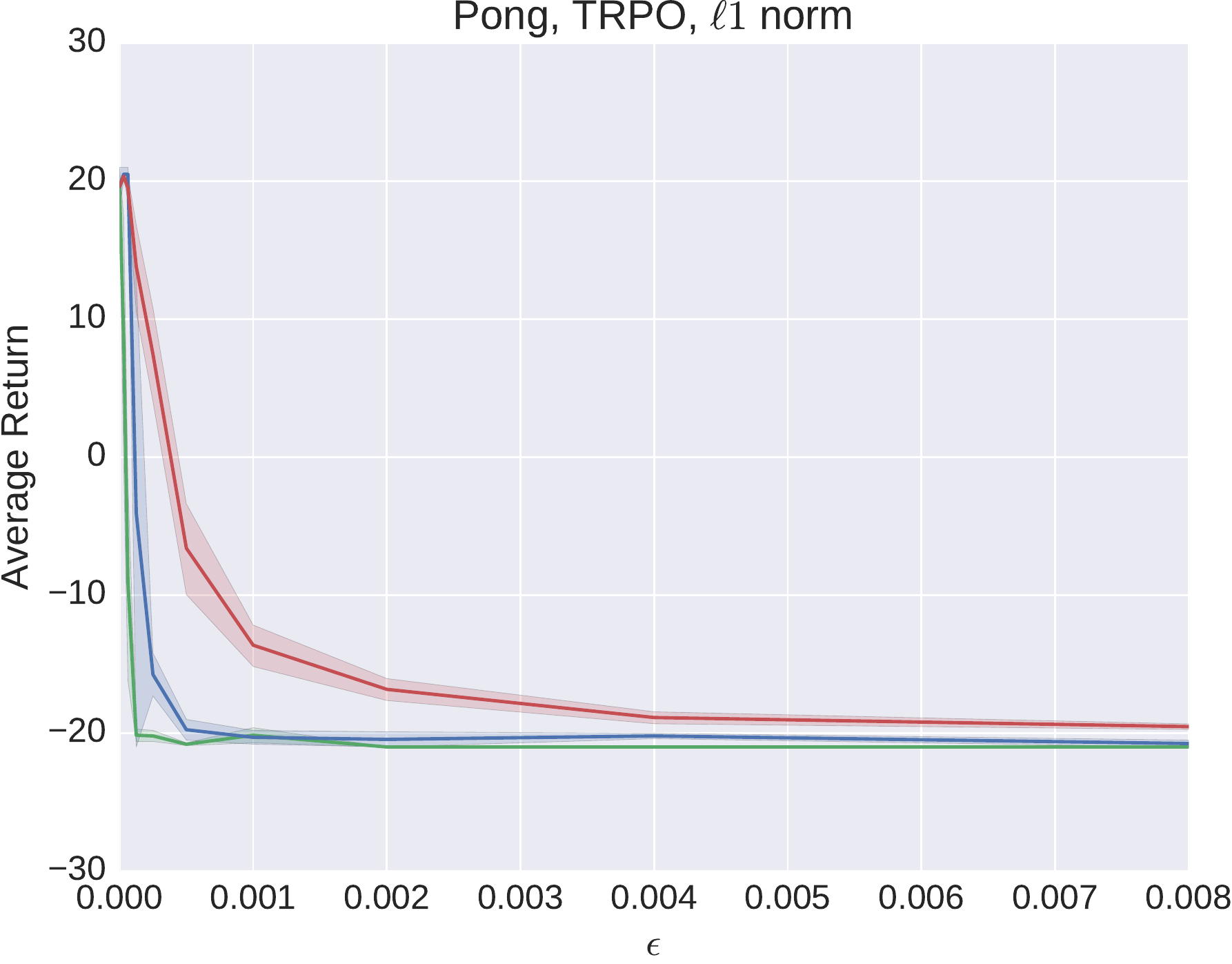}} &
\subfloat{\includegraphics[width = 1.33in,natwidth=521,natheight=402]{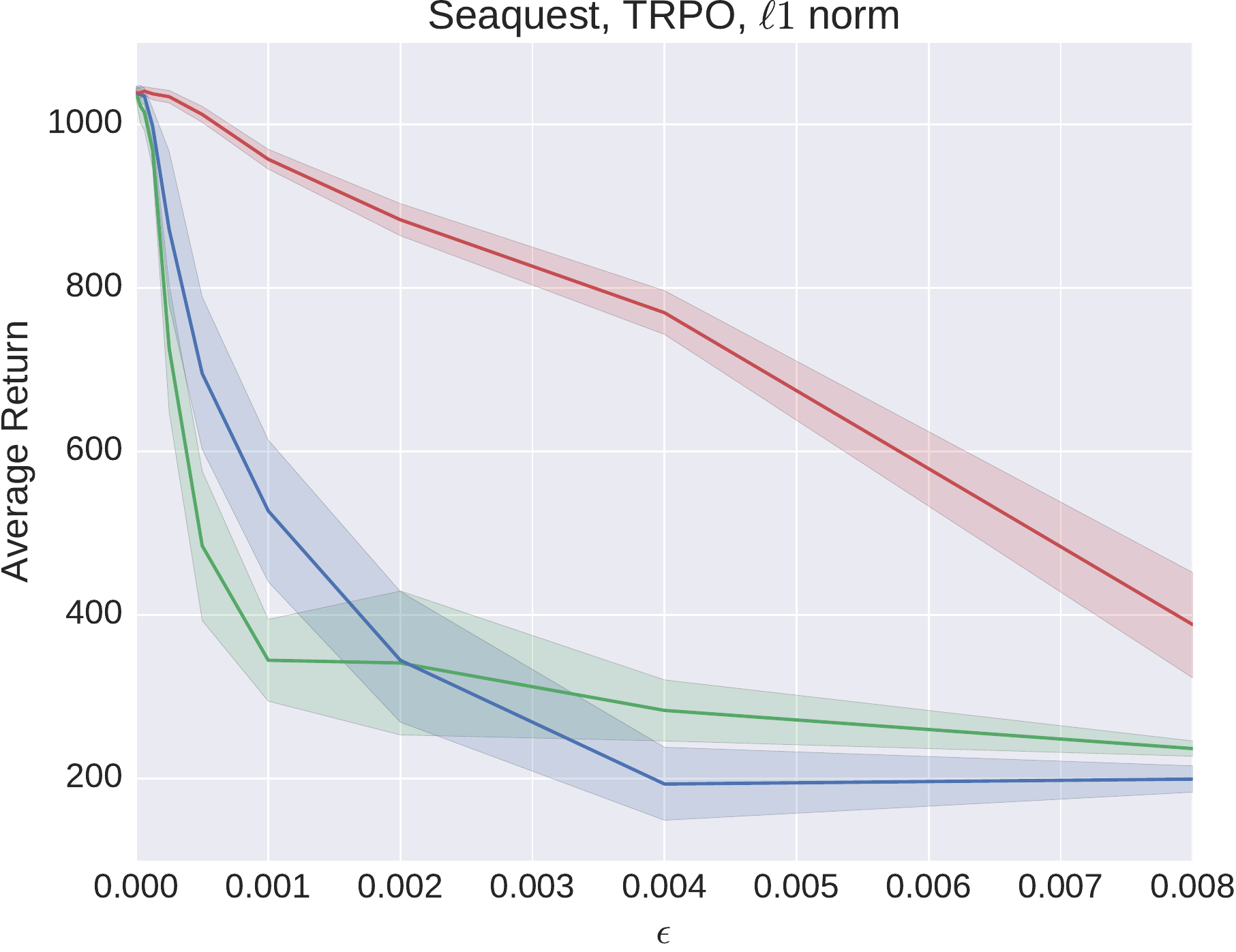}} &
\subfloat{\includegraphics[width = 1.33in,natwidth=521,natheight=402]{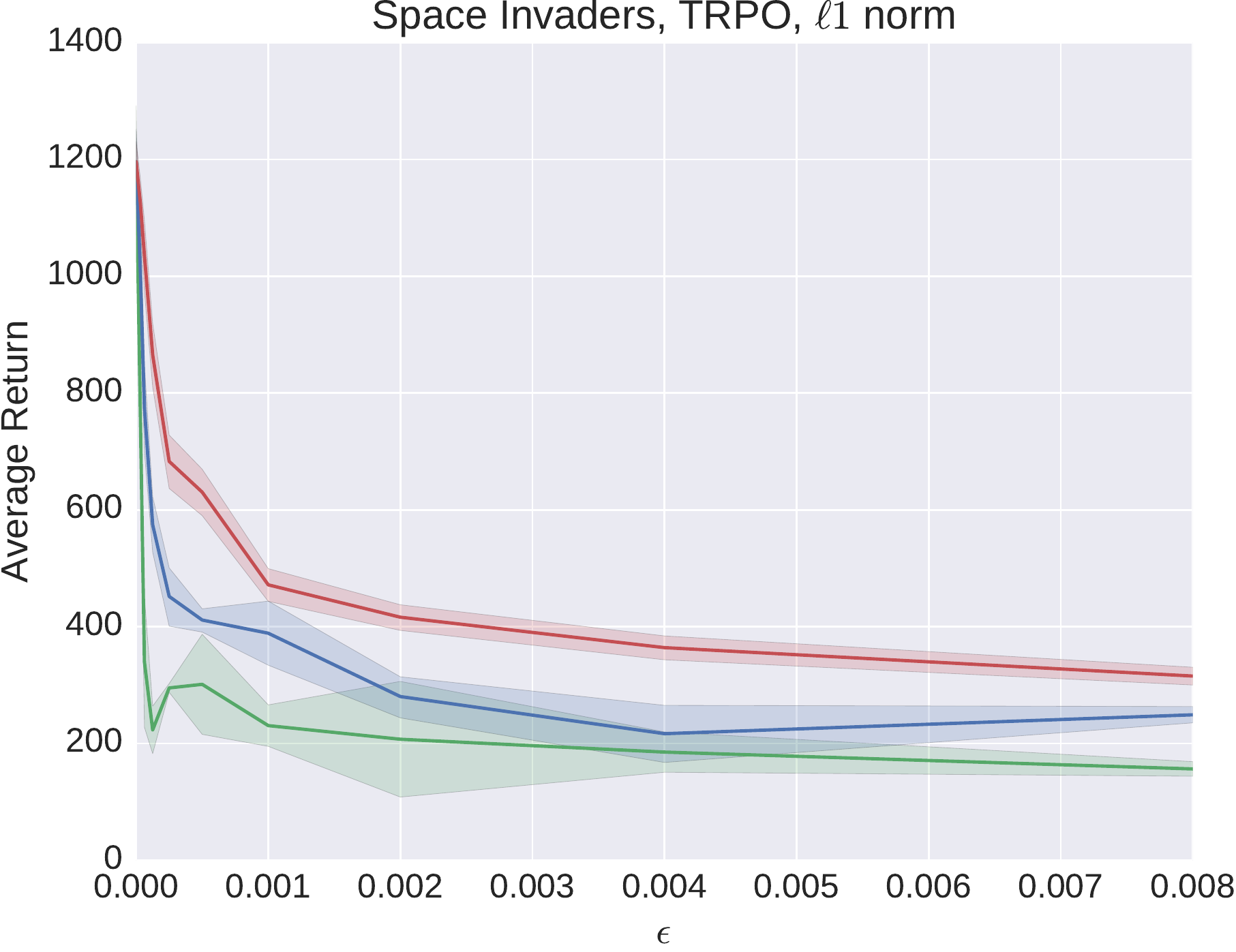}} \\
\addlinespace[-2ex]
\subfloat{\includegraphics[width = 1.33in,natwidth=521,natheight=402]{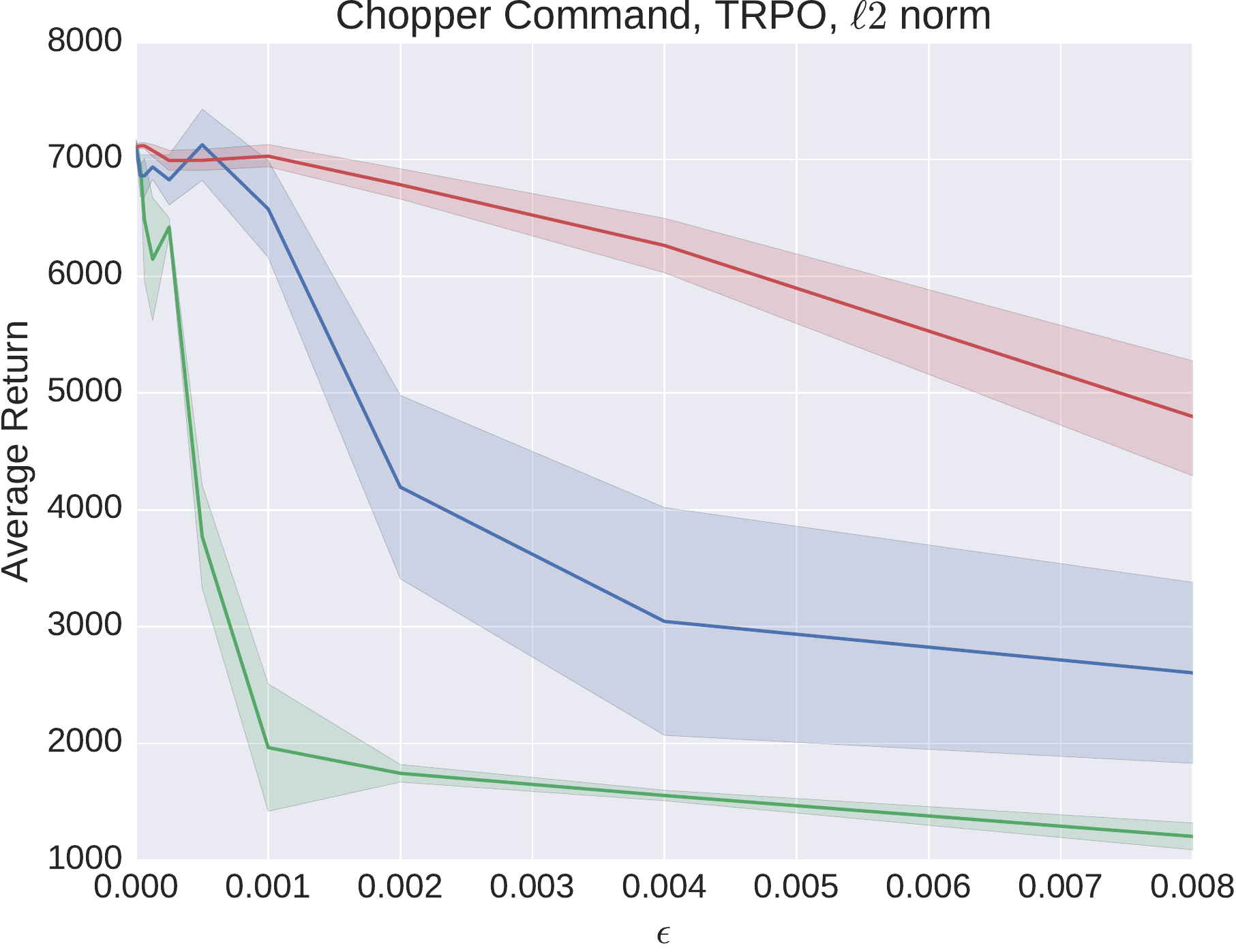}} &
\subfloat{\includegraphics[width = 1.33in,natwidth=514,natheight=402]{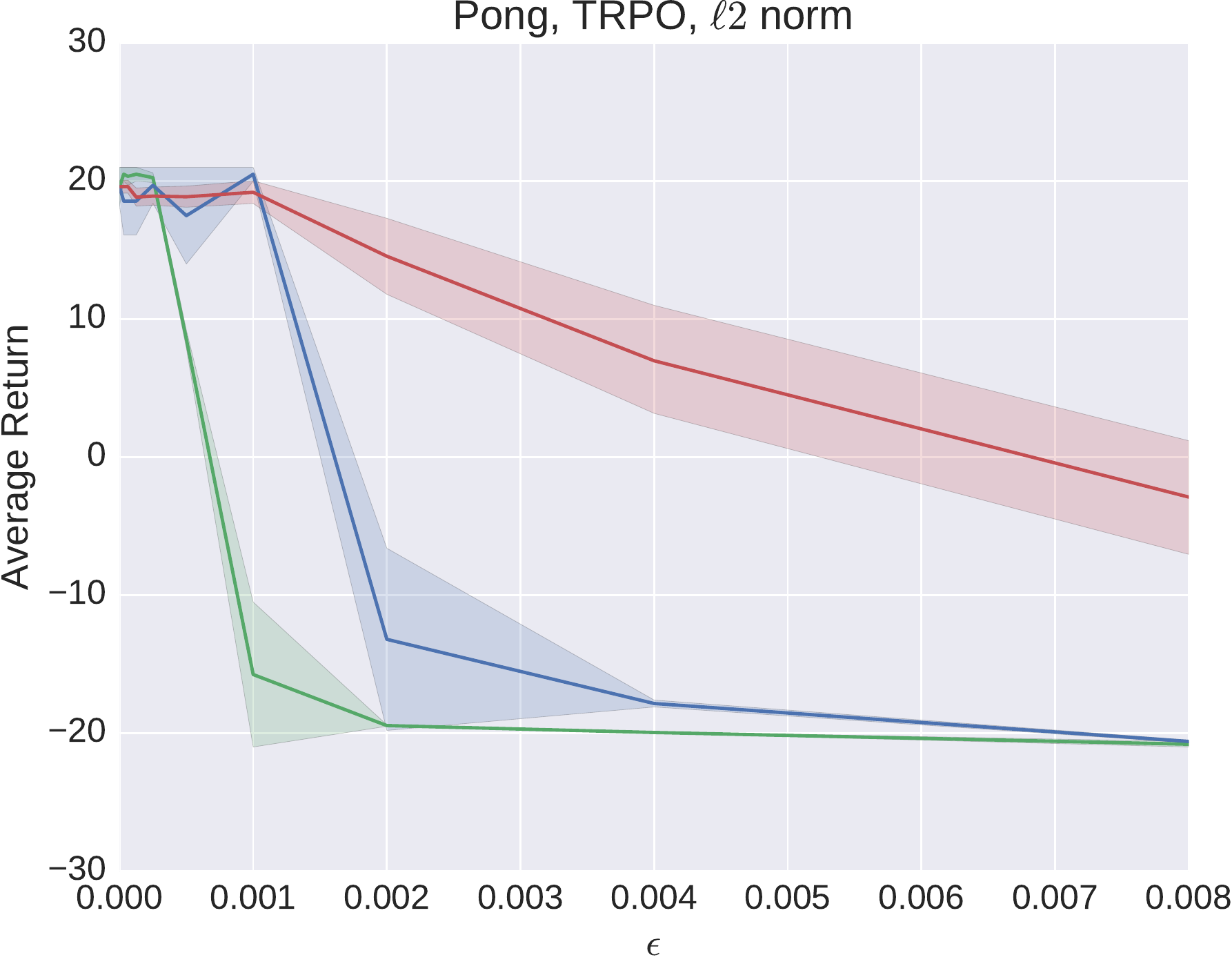}} &
\subfloat{\includegraphics[width = 1.33in,natwidth=521,natheight=402]{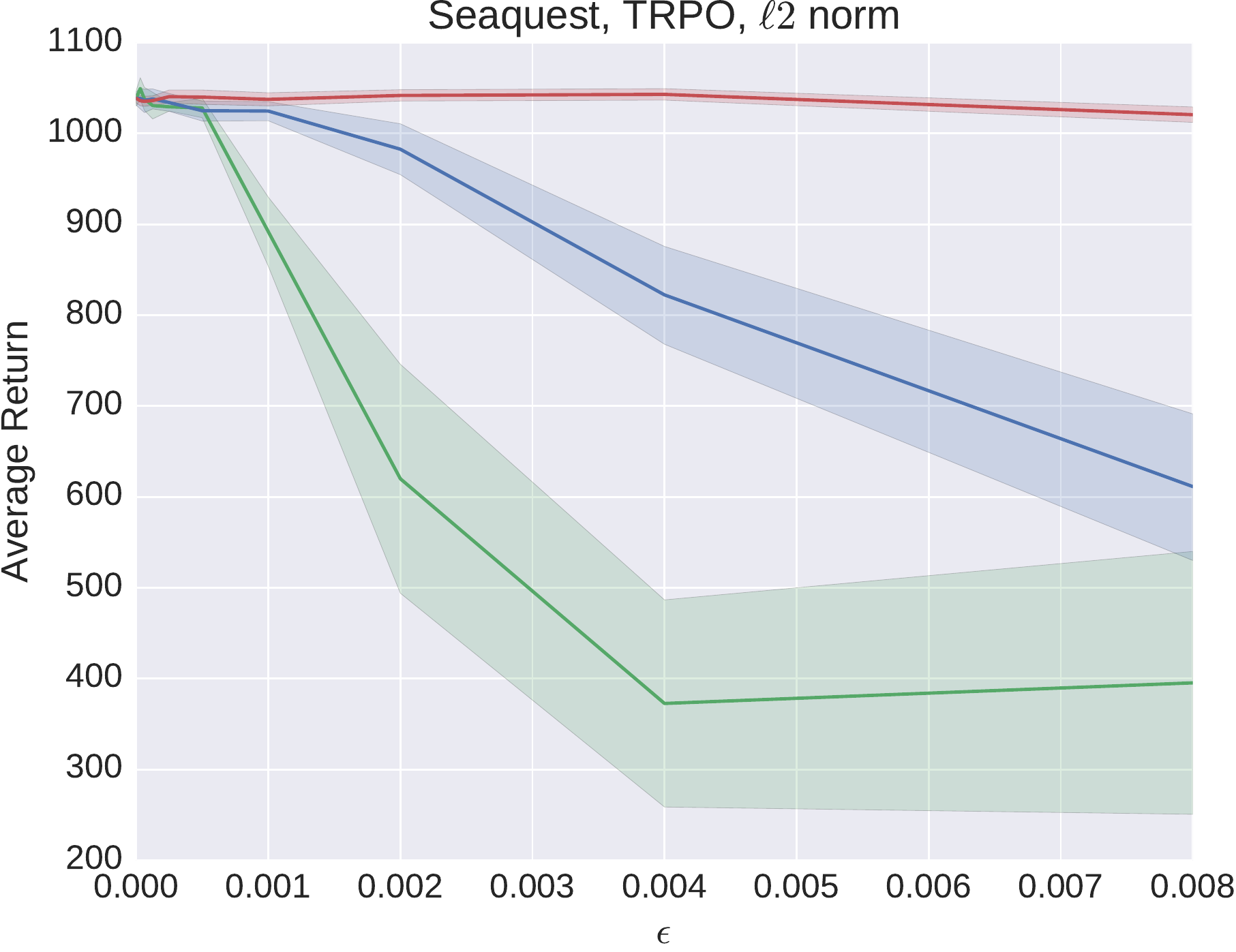}} &
\subfloat{\includegraphics[width = 1.33in,natwidth=521,natheight=402]{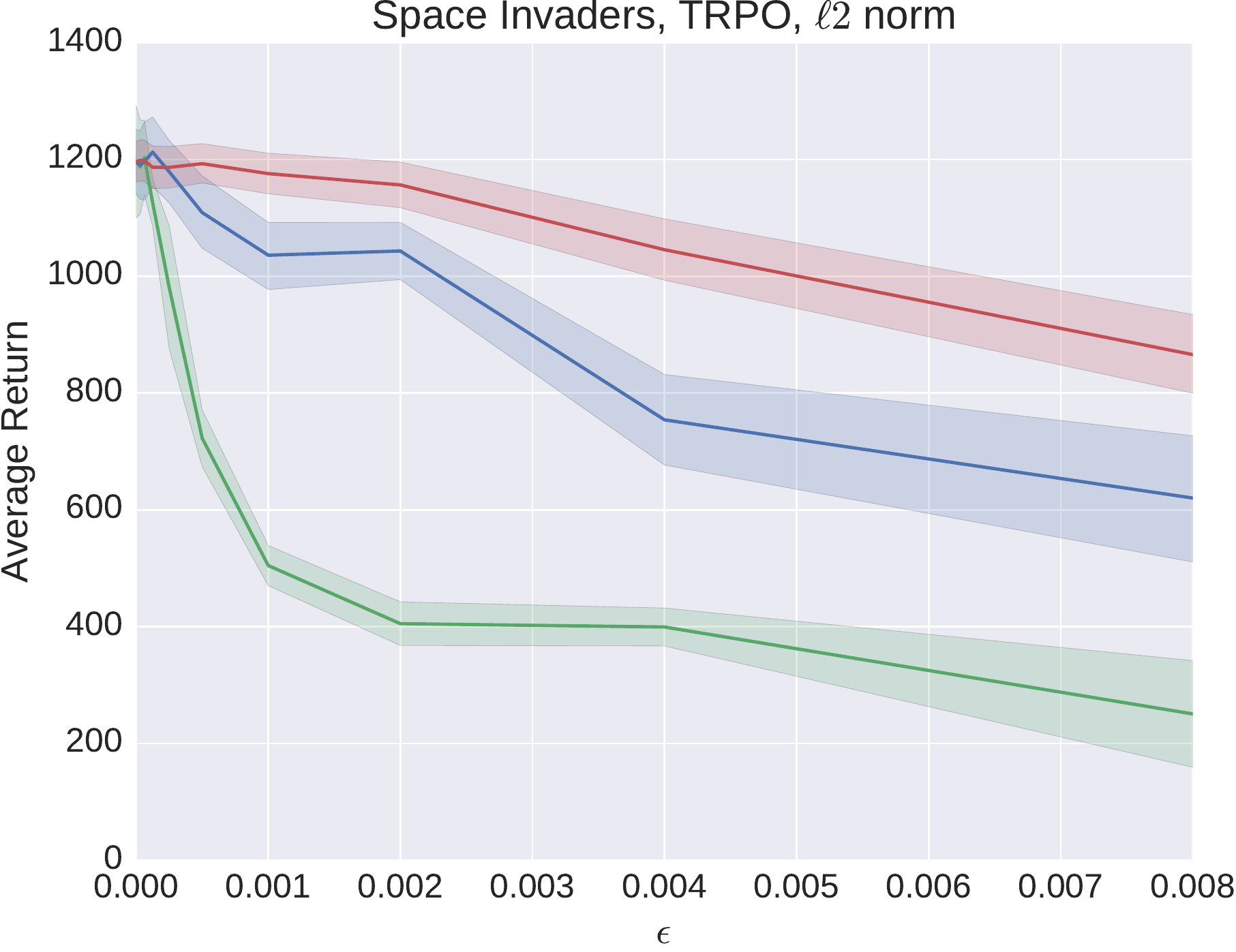}} 
\end{tabular}
\caption{Transferability of adversarial inputs for policies trained with TRPO. Type of transfer:
\crule[plot-algo]{0.3cm}{0.3cm} algorithm
\crule[plot-policy]{0.3cm}{0.3cm} policy
\crule[plot-none]{0.3cm}{0.3cm} none}
\label{fig:transfer-TRPO}
\end{figure*}

\begin{figure*}[t!]
\centering
\begin{tabular}{cccc}
\subfloat{\includegraphics[width = 1.33in,natwidth=521,natheight=402]{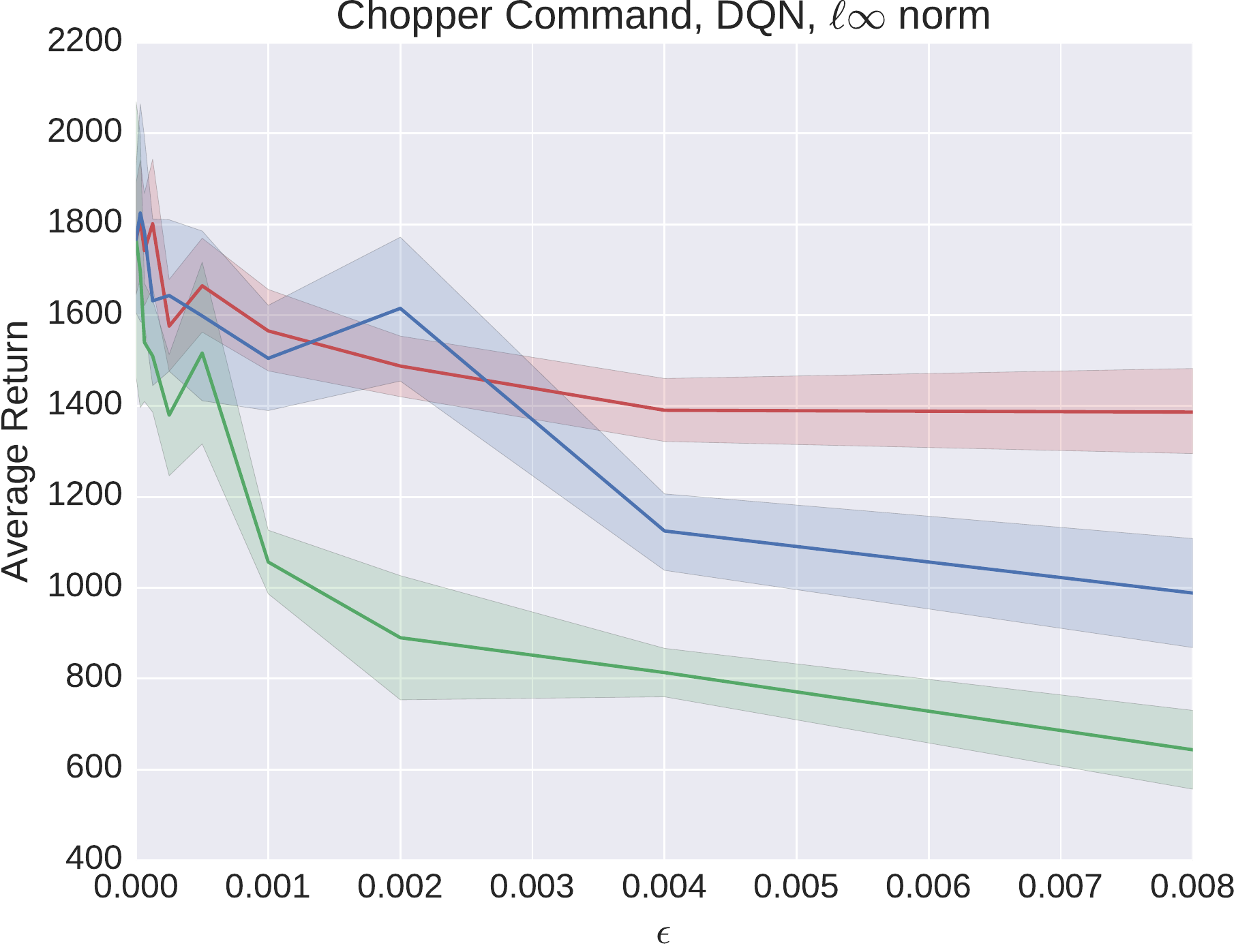}} &
\subfloat{\includegraphics[width = 1.33in,natwidth=514,natheight=402]{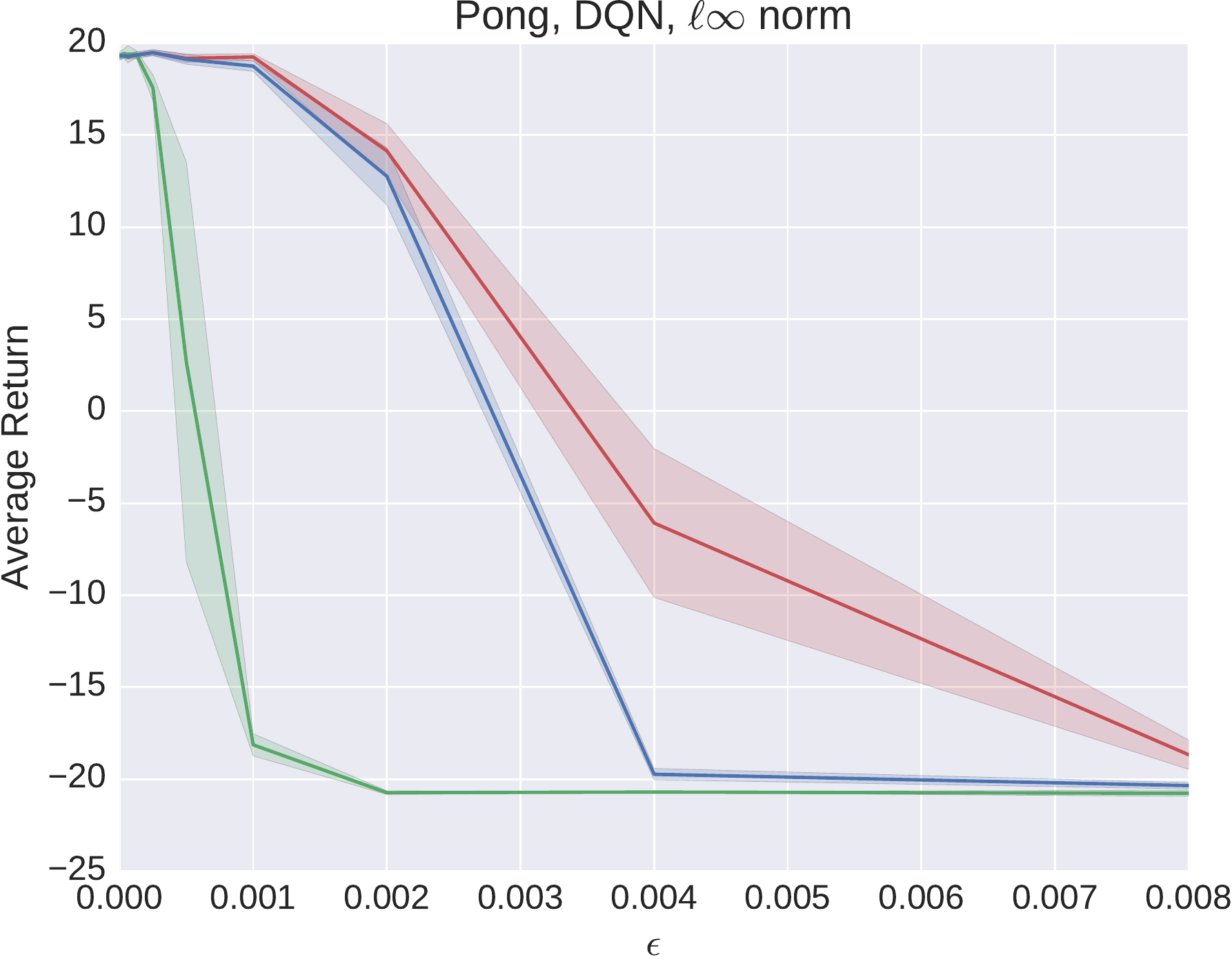}} &
\subfloat{\includegraphics[width = 1.33in,natwidth=521,natheight=402]{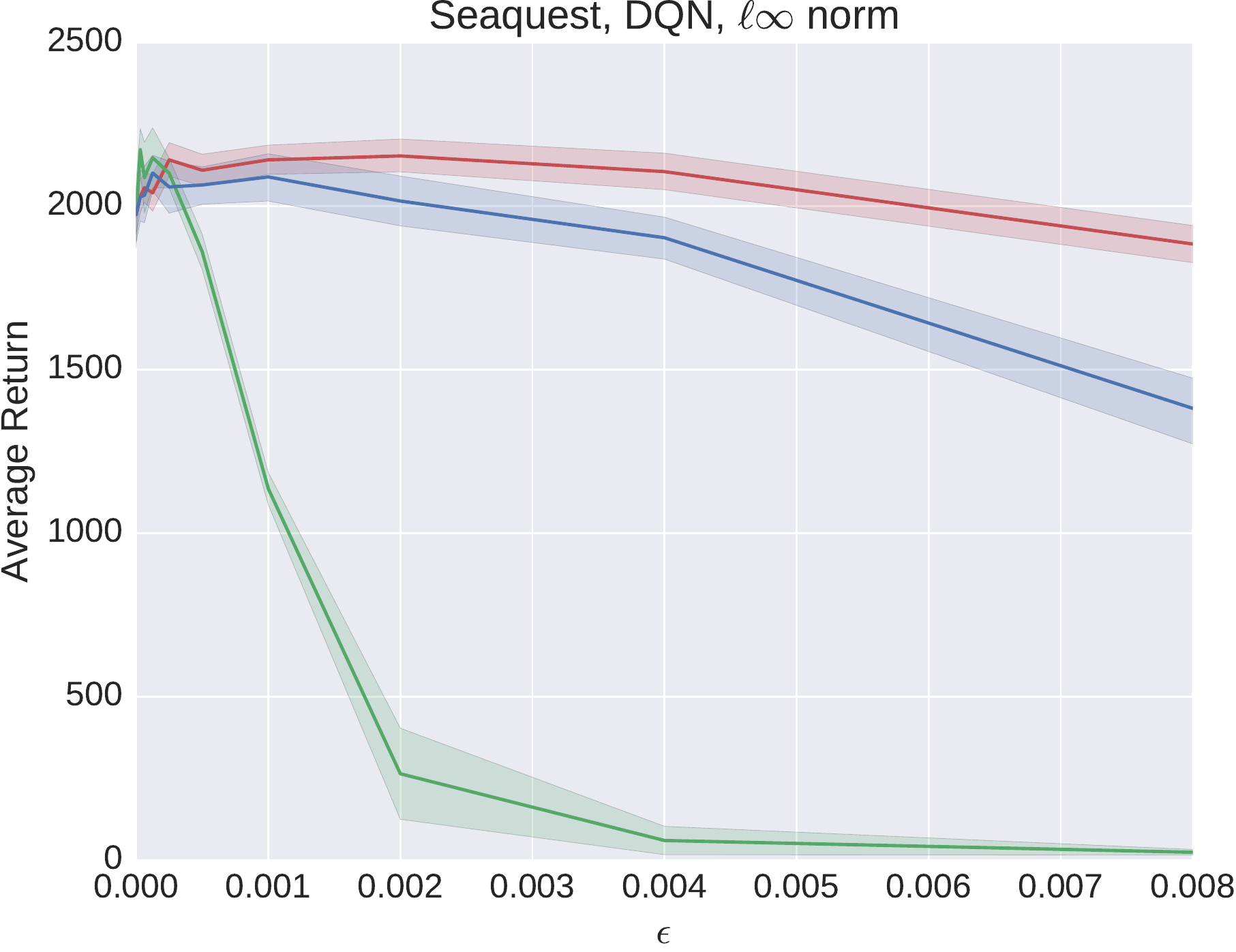}} &
\subfloat{\includegraphics[width = 1.33in,natwidth=513,natheight=402]{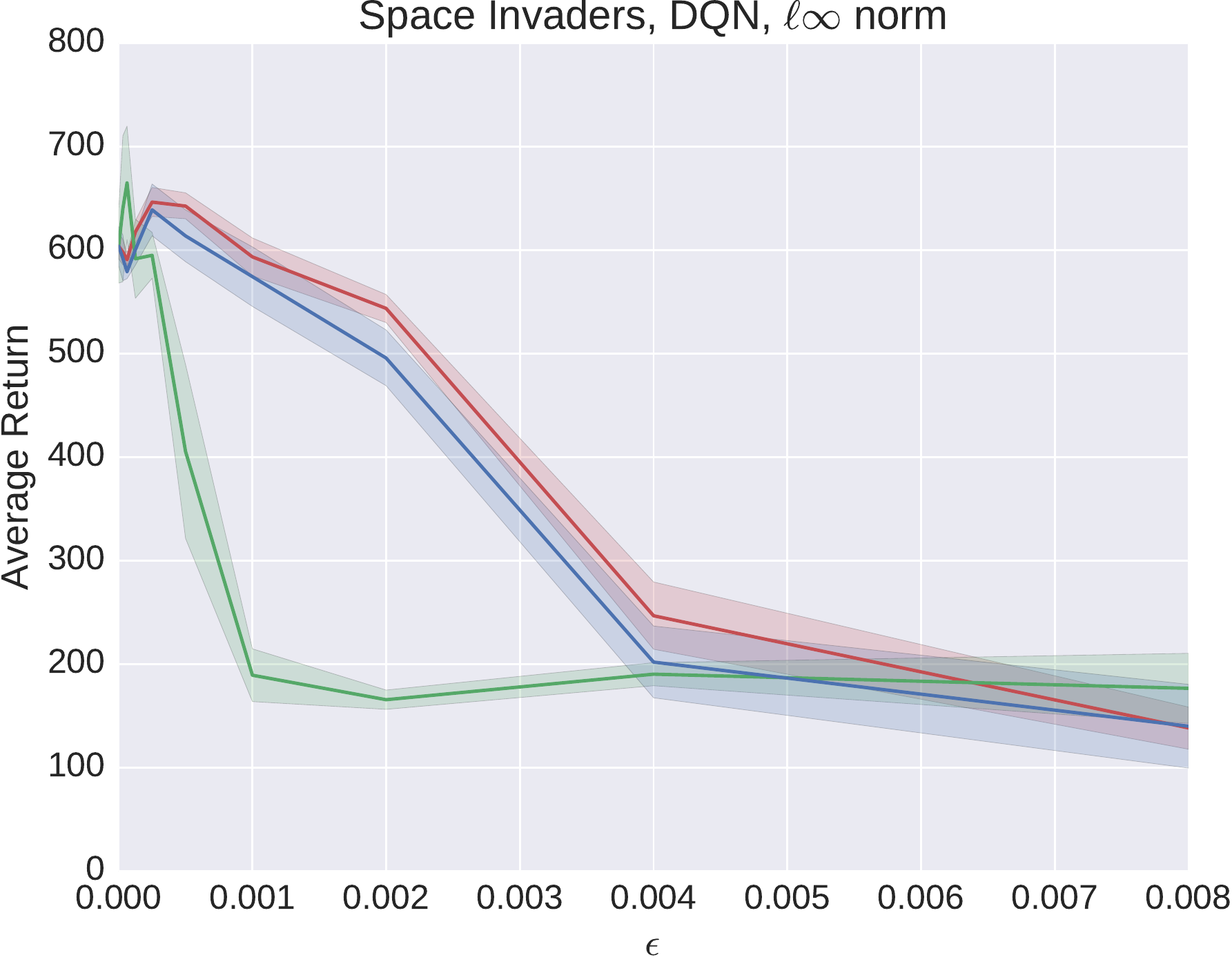}} \\
\addlinespace[-2ex]
\subfloat{\includegraphics[width = 1.33in,natwidth=521,natheight=402]{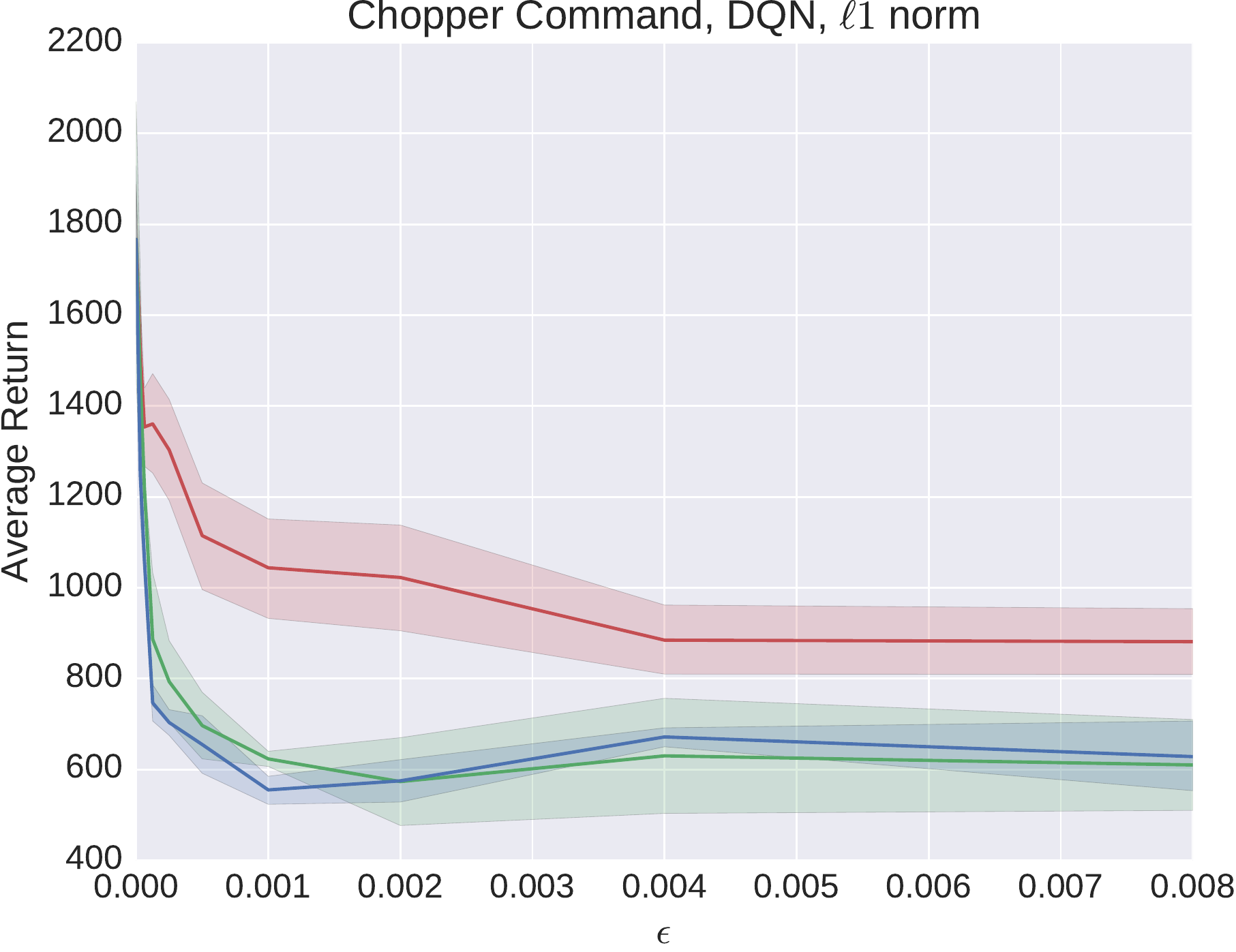}} &
\subfloat{\includegraphics[width = 1.33in,natwidth=514,natheight=402]{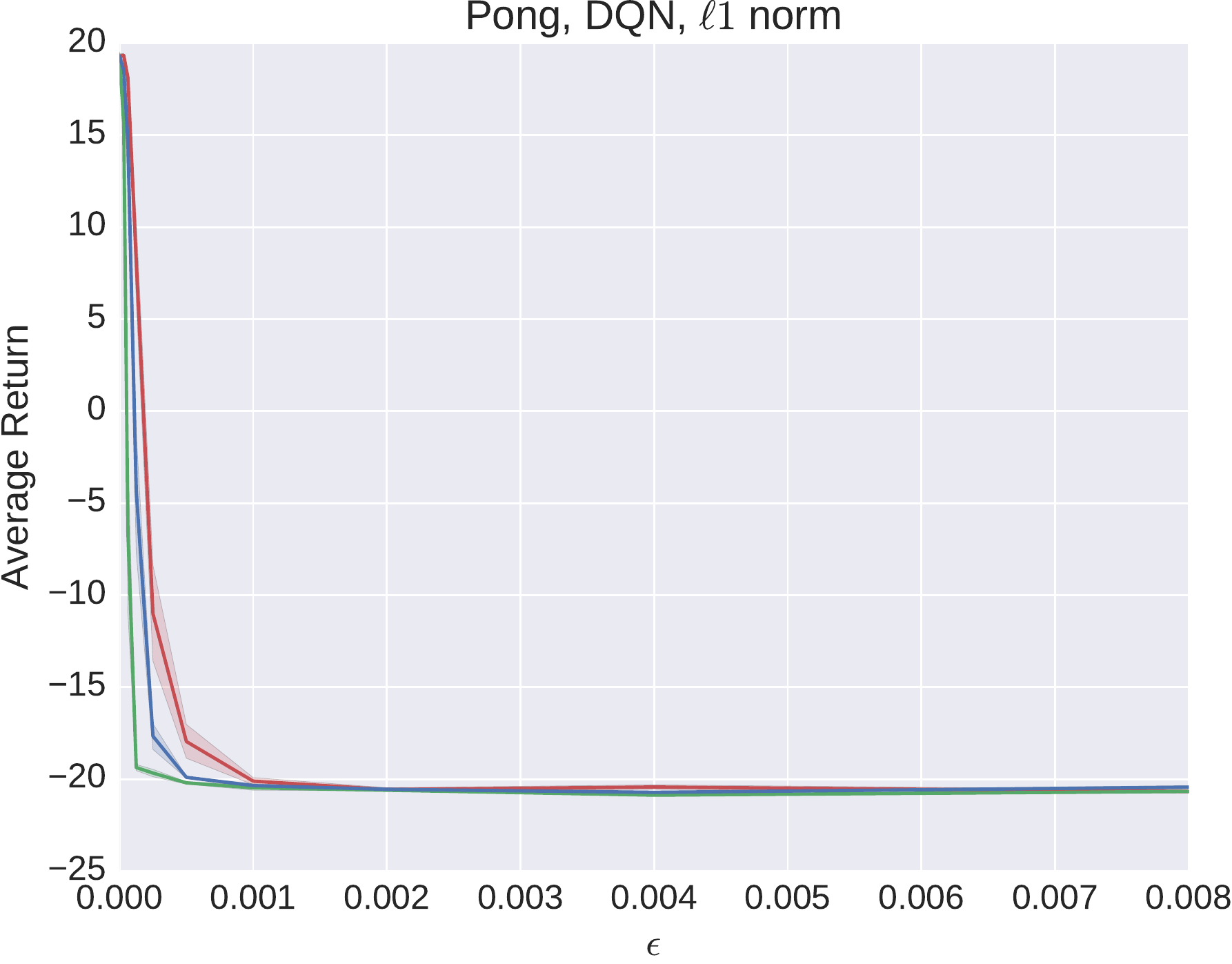}} &
\subfloat{\includegraphics[width = 1.33in,natwidth=521,natheight=402]{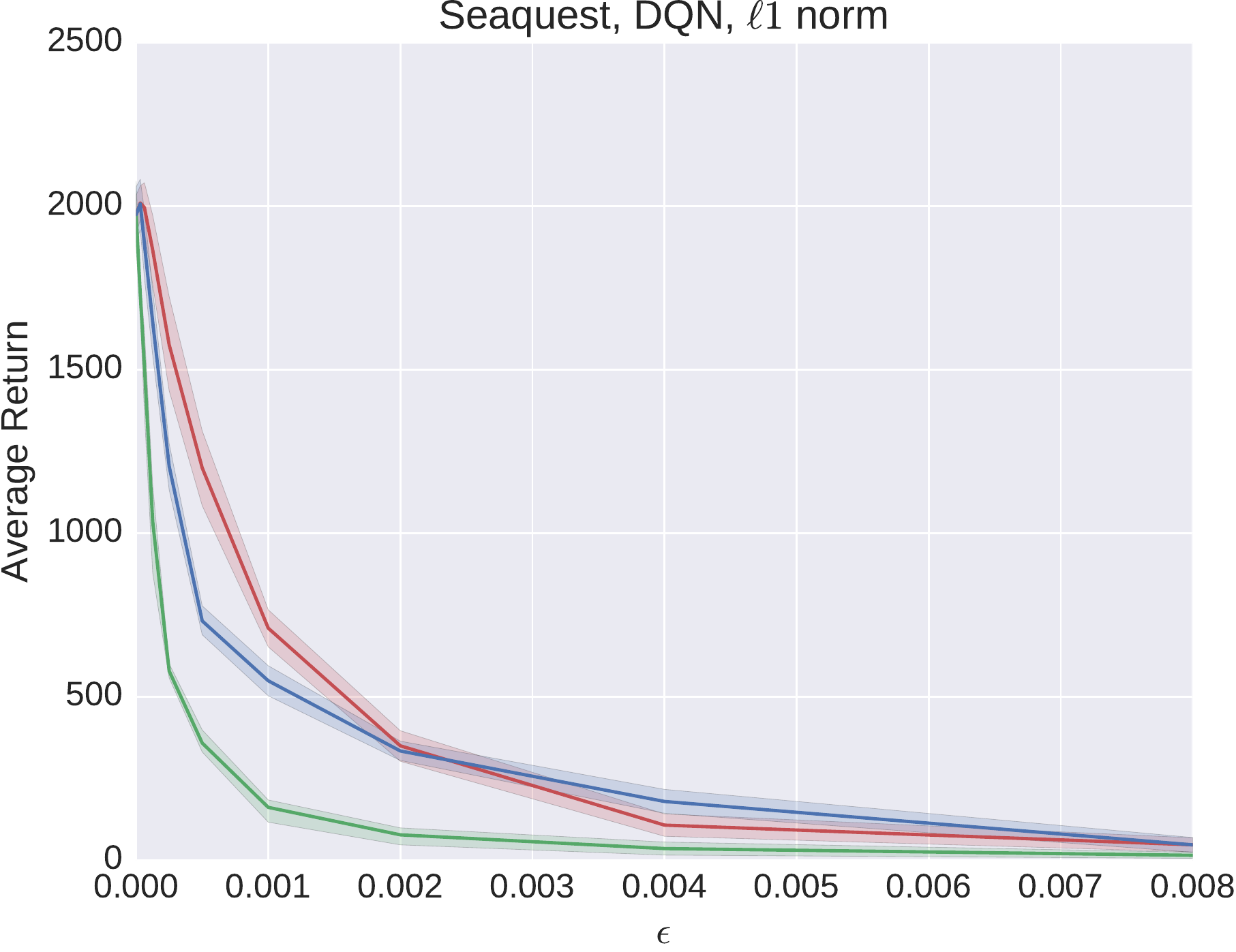}} &
\subfloat{\includegraphics[width = 1.33in,natwidth=513,natheight=402]{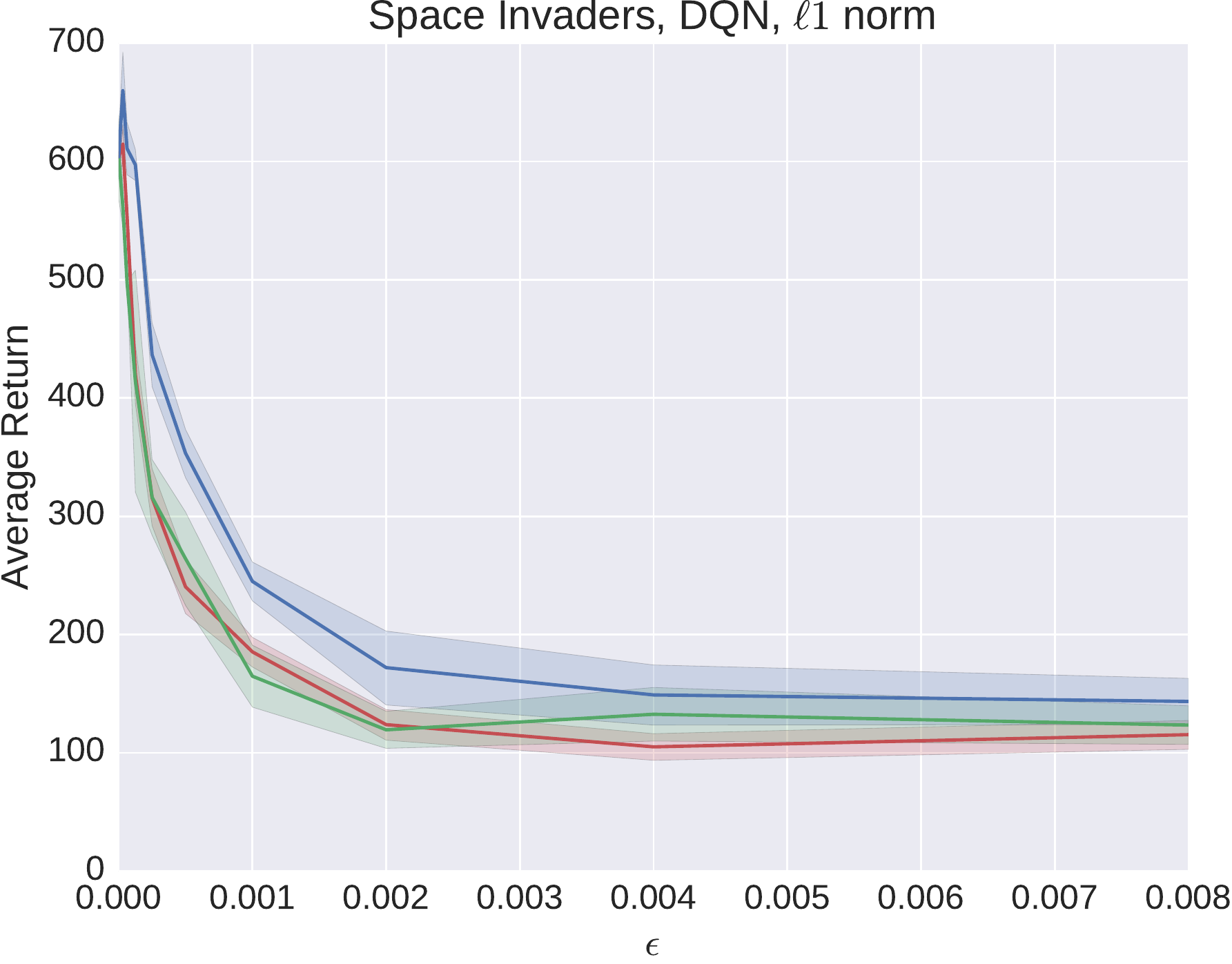}} \\
\addlinespace[-2ex]
\subfloat{\includegraphics[width = 1.33in,natwidth=521,natheight=402]{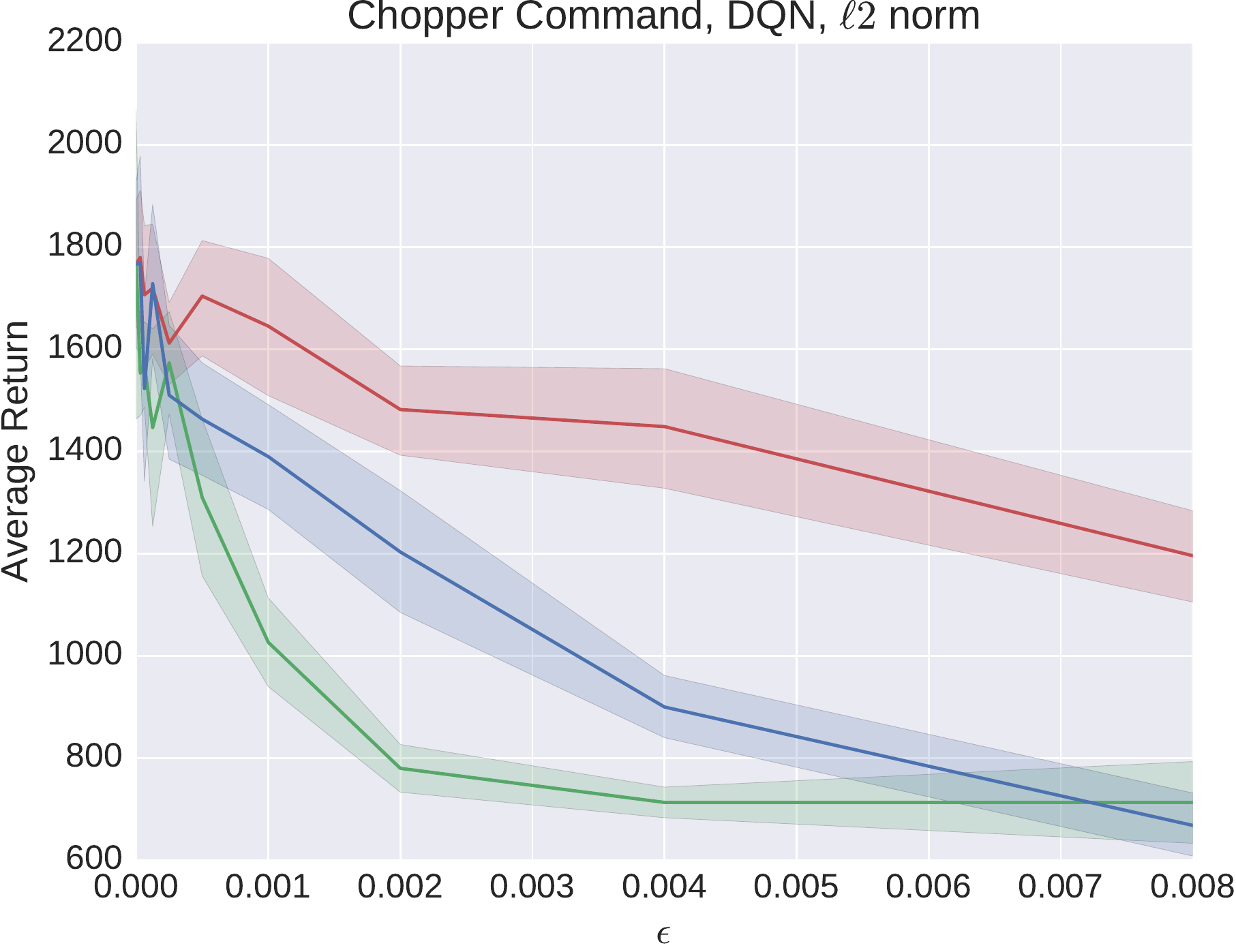}} &
\subfloat{\includegraphics[width = 1.33in,natwidth=514,natheight=402]{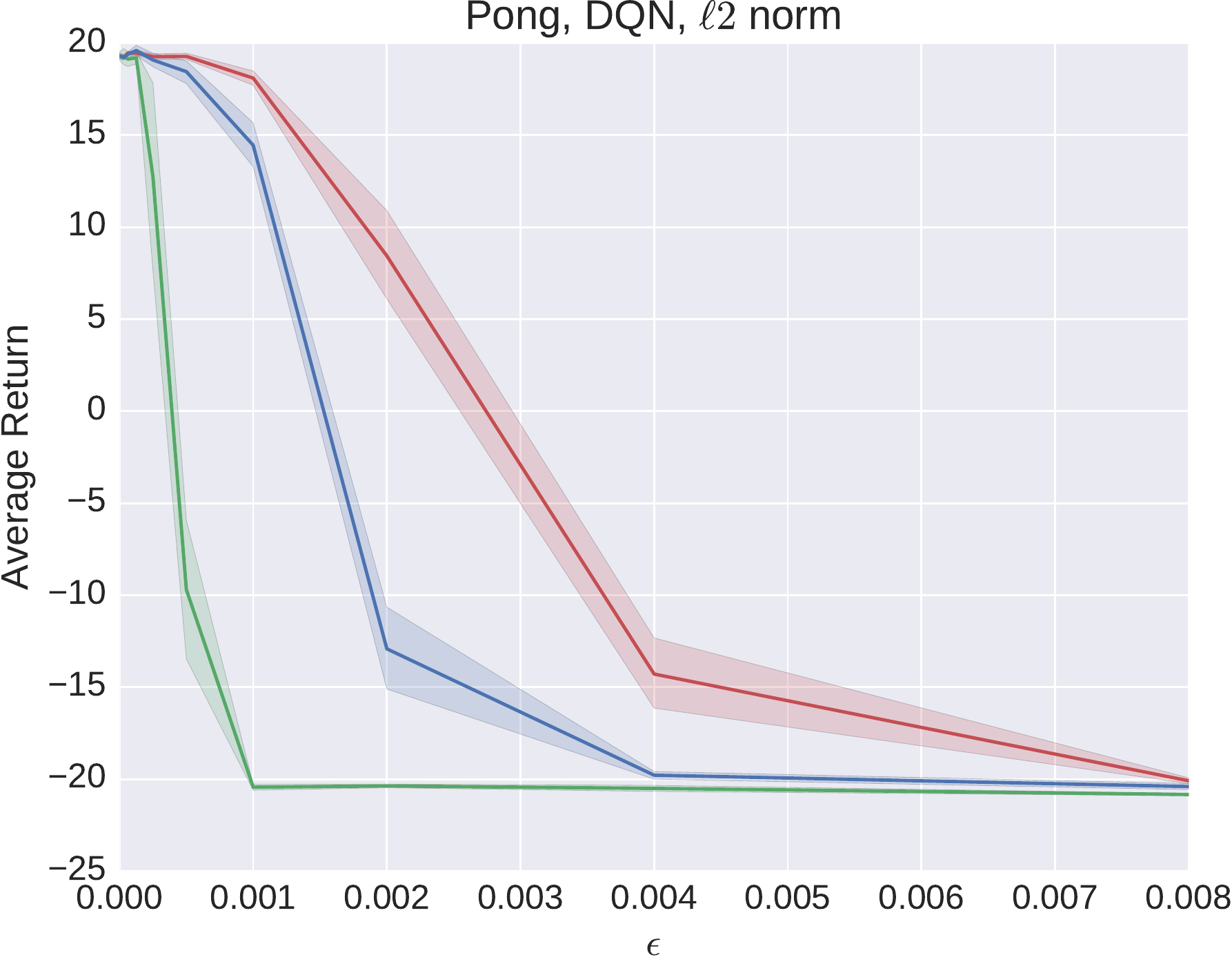}} &
\subfloat{\includegraphics[width = 1.33in,natwidth=521,natheight=402]{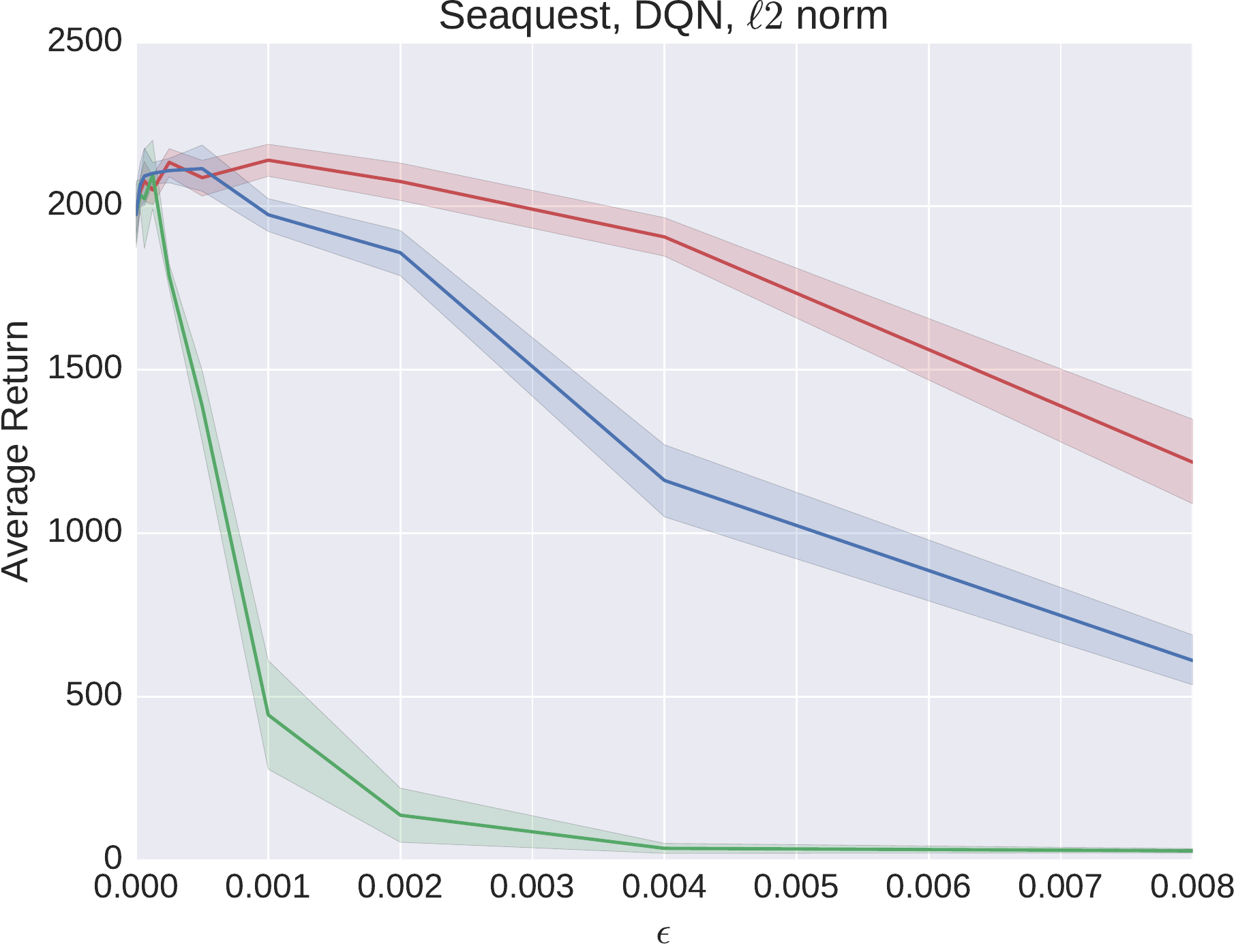}} &
\subfloat{\includegraphics[width = 1.33in,natwidth=513,natheight=402]{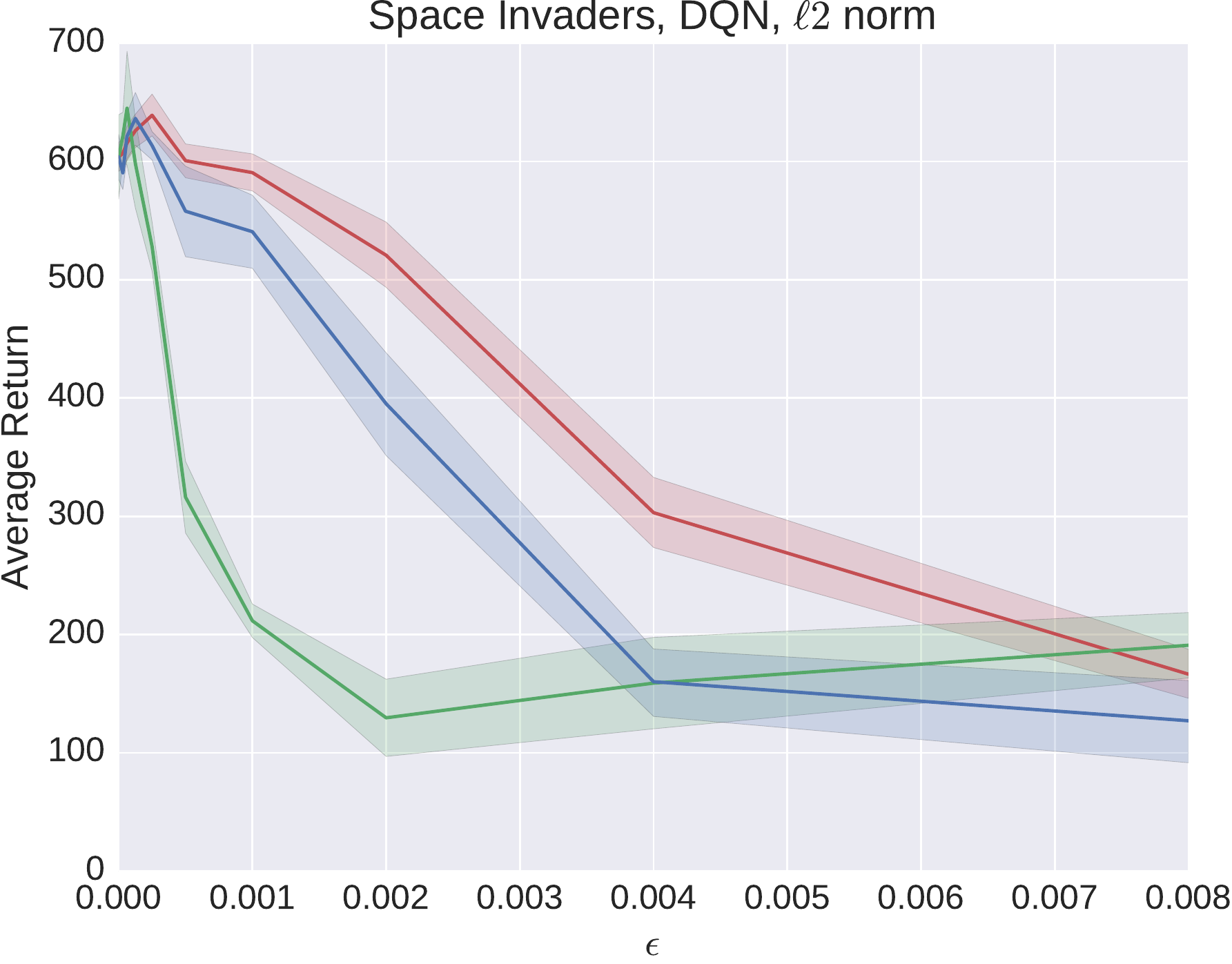}} 
\end{tabular}
\caption{Transferability of adversarial inputs for policies trained with DQN. Type of transfer:
\crule[plot-algo]{0.3cm}{0.3cm} algorithm
\crule[plot-policy]{0.3cm}{0.3cm} policy
\crule[plot-none]{0.3cm}{0.3cm} none}
\label{fig:transfer-DQN}
\end{figure*}

As one might expect, we find that the less the adversary knows about the target policy, the less effective the adversarial examples are (Fig.~\ref{fig:transfer-A3C},~\ref{fig:transfer-TRPO},~\ref{fig:transfer-DQN}). Transferability across algorithms is less effective at decreasing agent performance than transferability across policies, which is less effective than when the adversary does not need to rely on transferability (i.e., the adversary has full access to the target policy network). However, for most games, transferability across algorithms is still able to significantly decrease the agent's performance, especially for larger values of $\epsilon$.

Notably for $\ell 1$-norm adversaries, transferability across algorithms is nearly as effective as no transferability, for most game and algorithm combinations.

\setlength\tabcolsep{1.5pt}

\setlength\tabcolsep{6pt} 

\section{Discussion and Future Work}
This direction of work has significant implications for both online and real-world deployment of neural network policies. Our experiments show it is fairly easy to confuse such policies with computationally-efficient adversarial examples, even in black-box scenarios. Based on~\cite{Kurakin_2016}, it is possible that these adversarial perturbations could be applied to objects in the real world, for example adding strategically-placed paint to the surface of a road to confuse an autonomous car's lane-following policy.

Thus, an important direction of future work is developing defenses against adversarial attacks. This could involve adding adversarially-perturbed examples during training time (as in~\cite{Goodfellow_2015}), or it could involve detecting adversarial input at test time, to be able to deal with it appropriately.

\bibliographystyle{abbrv}
\bibliography{references}

\begin{thebibliography}{10}

\bibitem{Abbeel_2004}
P.~Abbeel and A.~Y. Ng.
\newblock Apprenticeship learning via inverse reinforcement learning.
\newblock In {\em Proceedings of the Twenty-First International Conference on
  Machine Learning}, 2004.

\bibitem{barreno2006can}
M.~Barreno, B.~Nelson, R.~Sears, A.~D. Joseph, and J.~D. Tygar.
\newblock Can machine learning be secure?
\newblock In {\em Proceedings of the 2006 ACM Symposium on Information,
  Computer and Communications Security}, pages 16--25, 2006.

\bibitem{behzadan2017vulnerability}
V.~Behzadan and A.~Munir.
\newblock Vulnerability of deep reinforcement learning to policy induction
  attacks.
\newblock {\em arXiv preprint arXiv:1701.04143}, 2017.

\bibitem{Bellemare_2013}
M.~G. {Bellemare}, Y.~{Naddaf}, J.~{Veness}, and M.~{Bowling}.
\newblock The arcade learning environment: An evaluation platform for general
  agents.
\newblock {\em Journal of Artificial Intelligence Research}, 47:253--279, 06
  2013.

\bibitem{biggio2013evasion}
B.~Biggio, I.~Corona, D.~Maiorca, B.~Nelson, N.~{\v{S}}rndi{\'c}, P.~Laskov,
  G.~Giacinto, and F.~Roli.
\newblock Evasion attacks against machine learning at test time.
\newblock In {\em Machine Learning and Knowledge Discovery in Databases}, pages
  387--402, 2013.

\bibitem{biggio2012poisoning}
B.~Biggio, B.~Nelson, and L.~Pavel.
\newblock Poisoning attacks against support vector machines.
\newblock In {\em Proceedings of the Twenty-Ninth International Conference on
  Machine Learning}, 2012.

\bibitem{Bojarski_2016}
M.~Bojarski, D.~D. Testa, D.~Dworakowski, B.~Firner, B.~Flepp, P.~Goyal, L.~D.
  Jackel, M.~Monfort, U.~Muller, J.~Zhang, X.~Zhang, J.~Zhao, and K.~Zieba.
\newblock End to end learning for self-driving cars.
\newblock {\em arXiv preprint arXiv:1604.07316}, 2016.

\bibitem{Brockman_2016}
G.~Brockman, V.~Cheung, L.~Pettersson, J.~Schneider, J.~Schulman, J.~Tang, and
  W.~Zaremba.
\newblock Openai gym, 2016.

\bibitem{Duan_2016}
Y.~Duan, X.~Chen, R.~Houthooft, J.~Schulman, and P.~Abbeel.
\newblock Benchmarking deep reinforcement learning for continuous control.
\newblock In {\em Proceedings of the Thirty-Third International Conference on
  Machine Learning}, 2016.

\bibitem{Goodfellow_2015}
I.~J. Goodfellow, J.~Shlens, and C.~Szegedy.
\newblock Explaining and harnessing adversarial examples.
\newblock In {\em Proceedings of the Third International Conference on Learning
  Representations}, 2015.

\bibitem{Kurakin_2016}
A.~Kurakin, I.~Goodfellow, and S.~Bengio.
\newblock Adversarial examples in the physical world.
\newblock {\em arXiv preprint arXiv:1607.02533}, 2016.

\bibitem{kurakin2017adversarial}
A.~Kurakin, I.~Goodfellow, and S.~Bengio.
\newblock Adversarial machine learning at scale.
\newblock {\em Proceedings of the Fifth International Conference on Learning
  Representations}, 2017.

\bibitem{Levine_2016}
S.~Levine, C.~Finn, T.~Darrell, and P.~Abbeel.
\newblock End-to-end training of deep visuomotor policies.
\newblock {\em Journal of Machine Learning Research}, 17(39):1--40, 2016.

\bibitem{Lillicrap_2016}
T.~P. Lillicrap, J.~J. Hunt, A.~Pritzel, N.~Heess, T.~Erez, Y.~Tassa,
  D.~Silver, and D.~Wierstra.
\newblock Continuous control with deep reinforcement learning.
\newblock In {\em Proceedings of the Fourth International Conference on
  Learning Representations}, 2016.

\bibitem{Mnih_2013}
V.~Mnih, K.~Kavukcuoglu, D.~Silver, A.~Graves, I.~Antonoglou, D.~Wierstra, and
  M.~Riedmiller.
\newblock Playing atari with deep reinforcement learning.
\newblock In {\em NIPS Workshop on Deep Learning}, 2013.

\bibitem{Mnih_2016}
V.~Mnih, A.~Puigdomenech~Badia, M.~Mirza, A.~Graves, T.~P. Lillicrap,
  T.~Harley, D.~Silver, and K.~Kavukcuoglu.
\newblock Asynchronous methods for deep reinforcement learning.
\newblock In {\em Proceedings of the Thirty-Third International Conference on
  Machine Learning}, 2016.

\bibitem{papernot2016practical}
N.~Papernot, P.~McDaniel, I.~Goodfellow, S.~Jha, Z.~B. Celik, and A.~Swami.
\newblock Practical black-box attacks against deep learning systems using
  adversarial examples.
\newblock {\em arXiv preprint arXiv:1602.02697}, 2016.

\bibitem{papernot2016towards}
N.~Papernot, P.~McDaniel, A.~Sinha, and M.~Wellman.
\newblock Towards the science of security and privacy in machine learning.
\newblock {\em arXiv preprint arXiv:1611.03814}, 2016.

\bibitem{Schulman_2015}
J.~Schulman, S.~Levine, P.~Moritz, M.~I. Jordan, and P.~Abbeel.
\newblock Trust region policy optimization.
\newblock In {\em Proceedings of the Thirty-Second International Conference on
  Machine Learning}, 2015.

\bibitem{sharif2016accessorize}
M.~Sharif, S.~Bhagavatula, L.~Bauer, and M.~K. Reiter.
\newblock Accessorize to a crime: Real and stealthy attacks on state-of-the-art
  face recognition.
\newblock In {\em Proceedings of the 2016 ACM SIGSAC Conference on Computer and
  Communications Security}, pages 1528--1540, 2016.

\bibitem{Silver_2016}
D.~Silver, A.~Huang, C.~J. Maddison, A.~Guez, L.~Sifre, G.~van~den Driessche,
  J.~Schrittwieser, I.~Antonoglou, V.~Panneershelvam, M.~Lanctot, S.~Dieleman,
  D.~Grewe, J.~Nham, N.~Kalchbrenner, I.~Sutskever, T.~Lillicrap, M.~Leach,
  K.~Kavukcuoglu, T.~Graepel, and D.~Hassabis.
\newblock Mastering the game of go with deep neural networks and tree search.
\newblock {\em Nature}, 529:484--503, 2016.

\bibitem{szegedy2013intriguing}
C.~Szegedy, W.~Zaremba, I.~Sutskever, J.~Bruna, D.~Erhan, I.~Goodfellow, and
  R.~Fergus.
\newblock Intriguing properties of neural networks.
\newblock In {\em Proceedings of the Second International Conference on
  Learning Representations}, 2014.

\end{thebibliography}

\appendix
\section{Experimental Setup}
\label{sec:exp_setup}
We set up our experiments within the rllab~\cite{Duan_2016} framework. We use a parallelized version of the rllab implementation of TRPO, and integrate outside implementations of DQN\footnote{\url{github.com/spragunr/deep\_q\_rl}} and A3C\footnote{\url{github.com/muupan/async-rl}}. We use OpenAI Gym environments~\cite{Brockman_2016} as the interface to the Arcade Learning Environment~\cite{Bellemare_2013}.

The policies use the network architecture from~\cite{Mnih_2013}: a convolutional layer with 16 filters of size $8\times8$ with a stride of 4, followed by a convolutional layer with 32 filters of size $4\times4$ with a stride of 2. The last layer is a fully-connected layer with 256 hidden units. All hidden layers are followed by a rectified nonlinearity.

For all games, we set the frame skip to 4 as in~\cite{Mnih_2013}. The frame skip specifies the number of times the agent's chosen action is repeated.

\subsection{Training}
We trained policies with TRPO and A3C on Amazon EC2 c4.8xlarge machines. For each policy, we ran TRPO for 2,000 iterations of 100,000 steps each, which took 1.5 to 2 days. We set the bound on the KL divergence to 0.01, as in~\cite{Schulman_2015}.

For A3C, we used 18 actor-learner threads and a learning rate of 0.0004. As in~\cite{Mnih_2016}, we use an entropy regularization weight of 0.01, use RMSProp for optimization with a decay factor of 0.99, update the policy and value networks every 5 time steps, and share all weights except the output layer between the policy and value networks. For each policy, we ran A3C for 200 iterations of 1,000,000 steps each, which took 1.5 to 2 days.

For DQN, we trained policies on Amazon EC2 p2.xlarge machines. We used 100,000 steps per epoch and trained for two days.

\end{document}